\documentclass[11pt]{article}

\usepackage{acl}
\usepackage{times}
\usepackage{latexsym}
\usepackage[T1]{fontenc}
\usepackage[utf8]{inputenc}
\usepackage{newunicodechar}
\newunicodechar{ą}{\k{a}}
\newunicodechar{ă}{\u{a}}
\usepackage{microtype}
\usepackage{inconsolata}
\usepackage{graphicx}
\usepackage{booktabs}
\usepackage{amssymb}
\usepackage{multirow}
\usepackage{colortbl}
\usepackage{placeins}
\usepackage{float}
\usepackage{tcolorbox}

\title{Rethinking the Role of Tool Graph Memory in Multi-Turn Tool Use}

\author{
\textbf{Uiji Hwang\textsuperscript{1}},
\textbf{Richeng Xuan\textsuperscript{2}},
\textbf{Taeuk Kim\textsuperscript{1*}}
\\
\textsuperscript{1}Hanyang University,
\textsuperscript{2}Hunyuan, Tencent,
\\
\texttt{\{willpower, kimtaeuk\}@hanyang.ac.kr}, \texttt{richxuan@tencent.com}
}

\begin{document}
\maketitle
\begingroup
\renewcommand{\thefootnote}{\fnsymbol{footnote}}
\footnotetext{\textsuperscript{*}Corresponding author.}
\endgroup

\begin{abstract}

Large language model (LLM) agents solve complex tasks by calling external tools over multiple turns.
Recent work has proposed tool graph memory, which stores past experience as tool transitions to guide subsequent selection.
However, it remains unclear whether this component consistently improves agents' tool use or undermines it.
We evaluate tool graph memory on nine models and two benchmarks, finding that it seldom improves significantly over baselines, and that its gains, where present, cost over twice the LLM calls.
The memory's guidance is often incorrect, and random or no guidance retains most of the gain, implying that the benefit stems from extra computation rather than the memory itself.
% implying that the benefit stems from attempting more tool calls rather than the memory itself.
Moreover, even with the correct next tool suggested, many failures remain, such as giving up without a tool call.
Storing the same experience instead as textual insights outperforms tool graph memory with almost no extra LLM calls.
% This suggests that memory serves better as a soft prior than as a hard constraint on tool selection, leading to a more effective and generalizable solution.
This suggests that memory serves better as a soft prior than as a prescription of the next tool, leading to a more effective and generalizable solution.

\end{abstract}

\section{Introduction}
\label{sec:intro}

Large language models (LLMs) are increasingly used to support tool-use agents~\cite{schick2023toolformer,yao2023react,qin2024toolllm} that interact with users and tools over multiple turns~\cite{lu-etal-2025-toolsandbox,patil2025the}. % to perform complex tasks. 
As agents encounter similar situations across tasks, memory enables them to retain and reuse experience from previous interactions~\cite{er2026memtoolagentleveragingmemorytool,xiao2025toolmemenhancingmultimodalagents,li2026doesmemoryhelpmultitrajectory}. 

More specifically, recent work such as ToolNet~\cite{liu2024toolnetconnectinglargelanguage} and H-EPM~\cite{li2026experienceevolving} has proposed the concept of \textbf{tool graph memory}, which represents prior experience as tool transitions in a graph, providing step-level guidance for selecting the next tool.
However, it is still an open question whether such guidance consistently benefits agents across multi-turn interactions.

\begin{figure}[t]
\centering 
\includegraphics[width=\columnwidth]{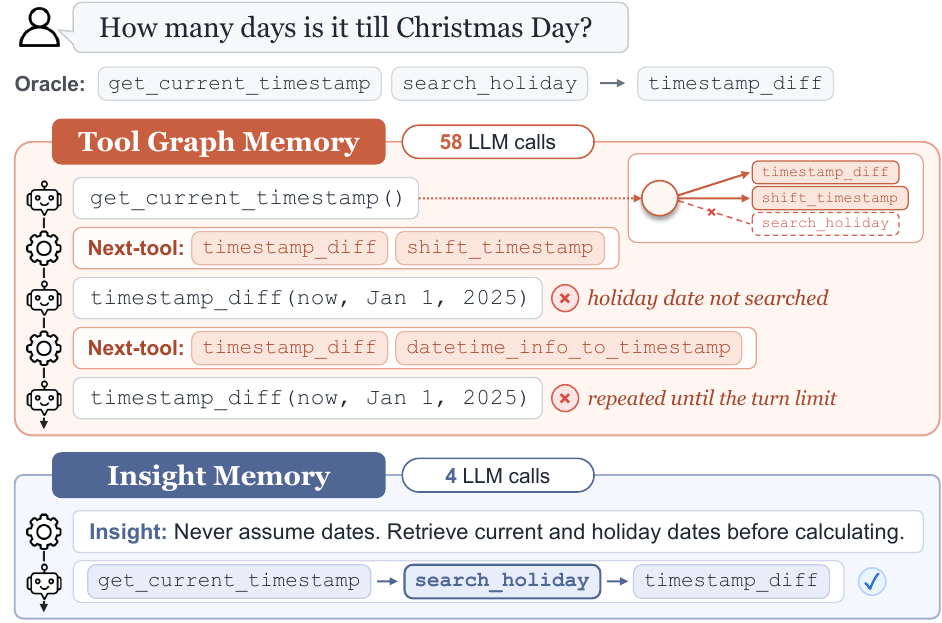}
% \caption{An example of tool graph memory and insight memory. The tool graph never suggests the prerequisite \texttt{set\_low\_battery\_mode\_status}; following the insight, the agent disables low battery mode first and succeeds.}
\caption{An example of tool graph memory and insight memory. The tool graph never suggests the prerequisite \texttt{search\_holiday}; following the insight, the agent retrieves the holiday date first and succeeds.}
\label{fig:overview}
\end{figure}
\begin{table*}[t]
\centering
\footnotesize
\setlength{\tabcolsep}{8.5pt}
\renewcommand{\arraystretch}{1.0}
\setlength{\aboverulesep}{1pt}
\setlength{\belowrulesep}{1pt}
\begin{tabular}{l cccc cccc}
\toprule
\multicolumn{1}{l}{\multirow{2}{*}{\textbf{Model}}} & \multicolumn{4}{c}{\textbf{ToolSandbox}} & \multicolumn{4}{c}{\textbf{BFCL-v4}} \\
\cmidrule(lr){2-5}\cmidrule(lr){6-9}
 & Base & ToolNet & H-EPM & Insight & Base & ToolNet & H-EPM & Insight \\
\midrule
Qwen3.5-2B & 0.420 & 0.428 & 0.436 & \textbf{0.488}\rlap{$^{+}$} & 0.133 & 0.044\rlap{$^{-}$} & 0.138 & \textbf{0.165} \\
Qwen3-8B & 0.500 & 0.507 & 0.476 & \textbf{0.556}\rlap{$^{+}$} & 0.171 & 0.129 & 0.179 & \textbf{0.242}\rlap{$^{+}$} \\
Qwen3.6-35B-A3B & 0.626 & 0.608 & 0.670 & \textbf{0.674} & 0.181 & 0.108\rlap{$^{-}$} & 0.156 & \textbf{0.206} \\
\addlinespace[0.7pt]
gemma-4-E4B-it & 0.588 & 0.547\rlap{$^{-}$} & 0.554 & \textbf{0.620} & 0.221 & 0.094\rlap{$^{-}$} & 0.252 & \textbf{0.285}\rlap{$^{+}$} \\
gemma-4-26B-A4B-it & 0.690 & 0.655\rlap{$^{-}$} & \textbf{0.705} & \textbf{0.705} & \textbf{0.566} & 0.369\rlap{$^{-}$} & 0.519 & 0.531 \\
\addlinespace[0.7pt]
Ministral-3-8B & 0.631 & 0.588\rlap{$^{-}$} & \textbf{0.672}\rlap{$^{+}$} & 0.539\rlap{$^{-}$} & 0.209 & 0.141\rlap{$^{-}$} & \textbf{0.328}\rlap{$^{+}$} & 0.306\rlap{$^{+}$} \\
Mistral-Small-3.2-24B & 0.545 & 0.557 & 0.603\rlap{$^{+}$} & \textbf{0.620}\rlap{$^{+}$} & 0.231 & 0.127\rlap{$^{-}$} & 0.250 & \textbf{0.338}\rlap{$^{+}$} \\
\addlinespace[0.7pt]
gpt-oss-20b & 0.579 & 0.573 & 0.620 & \textbf{0.630}\rlap{$^{+}$} & 0.325 & 0.152\rlap{$^{-}$} & 0.369 & \textbf{0.427}\rlap{$^{+}$} \\
gpt-oss-120b & 0.726 & 0.674\rlap{$^{-}$} & 0.713 & \textbf{0.734} & 0.494 & 0.275\rlap{$^{-}$} & \textbf{0.559}\rlap{$^{+}$} & 0.454 \\
\bottomrule
\end{tabular}
\caption{Performance of LLM agents on ToolSandbox and the multi-turn subset of BFCL-v4, averaged over three runs. The best method per benchmark is in \textbf{bold}. The plus ($+$) and minus $(-)$ signs indicate significant improvements and decrements over Base ($p<0.05$). Standard deviations and $p$-values are reported in Table~\ref{tab:main-full}.}
\label{tab:main}
\end{table*}

In this paper, we revisit the effectiveness of tool graph memory. 
First, we evaluate whether tool graph memory consistently improves over memory-free baselines across model families, sizes, and benchmarks, using repeated runs and paired significance tests.
Second, we quantify the additional computational cost it imposes, measured in LLM calls and tokens. 
Third, through manual analysis and controlled experiments, we examine the source of its gains, where present.
Finally, we identify the failure modes that next-tool suggestions alone fail to resolve and characterize the form of guidance required beyond such suggestions.

Surprisingly, we find that tool graph memory rarely yields a significant gain over the base agent across models and benchmarks, and that the gains of H-EPM require substantially more LLM calls and tokens.
Moreover, these gains do not stem from the memory itself, whose suggestions are often incorrect: replacing the suggested tools with randomly chosen ones, or removing the suggestions altogether, retains most of the improvement.
Instead, the improvement comes from the agent making more tool calls; indeed, a simple instruction encouraging the agent to keep calling tools yields comparable gains.

Furthermore, even with gold-standard tools injected as an upper bound, many failures persist, e.g., giving up without calling a necessary tool or invoking one before its prerequisites are met.
Next-tool suggestions thus cannot fully resolve them.
We hypothesize that the agent requires higher-level guidance on what to do in its situation, which \textbf{insight memory}---a form of semantic memory~\cite{sumers2024cognitive,liang2025aimeetsbrainmemory,du2026memoryautonomousllmagentsmechanisms} in natural language---can provide (see Figure~\ref{fig:overview}).
Insight memory recovers most of the scenarios where tool graph memory degrades and outperforms it in most settings with almost no extra LLM calls.
These results suggest that, in multi-turn tool use, agents benefit more from such situation-level cues than from direct tool suggestions.

\section{Limited Benefit of Tool Graph Memory}
\label{sec:revisit}
\subsection{Tool Graph Memory}
\label{sec:toolgraph}
Tool graph memory builds a graph of tool transitions from successful trajectories, as in Figure~\ref{fig:overview}. 

ToolNet~\cite{liu2024toolnetconnectinglargelanguage} represents tools as nodes and transitions as weighted, directed edges, where weights reflect how frequently each transition occurs. It presents the agent with only the successors of the previously invoked tool, using these weights as preference scores.

% In contrast, H-EPM~\cite{li2026experienceevolving} keeps the full tool set and stores on each edge a weight, \textcolor{red}{its procedural memory}, along with episodic summaries of the contexts in which the transition occurred.
In contrast, H-EPM~\cite{li2026experienceevolving} keeps the full tool set; its weighted tool transitions serve as procedural memory, and the summaries of the contexts in which each transition occurred, attached to the edges, as episodic memory.
At each step, the agent decides whether to summarize the current state.
If so, H-EPM suggests the two successor tools whose stored contexts best match the resulting summary; otherwise, the two with the highest transition weights.
We refer to the candidate tools provided by both methods as next-tool suggestions, or the guidance of tool graph memory.\footnote{We use H-EPM's inference-time memory variant, its strongest in our setting, not the reinforcement learning one.}

\subsection{Experimental Settings}
\label{sec:setup}
We compare agents augmented with tool graph memory against memory-free variants (Base).
Both are evaluated on two multi-turn tool-use benchmarks---ToolSandbox~\cite{lu-etal-2025-toolsandbox} and the corresponding subset of BFCL-v4~\cite{patil2025the}---with task performance reported on a 0--1 scale.
We employ nine open-weight models from the Qwen, Gemma, Mistral, and gpt-oss families, ranging from 2B to 120B parameters.
Each method is run three times, and we report the mean and standard deviation.
Details of the benchmarks and experimental setup are provided in Appendix~\ref{app:setup}.

\subsection{Experimental Results}
\label{sec:results}

\paragraph{Performance.}
Table~\ref{tab:main} shows that neither ToolNet nor H-EPM consistently improves over Base.
ToolNet underperforms Base in most cases, including all nine models on BFCL-v4, with a statistically significant gap in 12 ($p<0.05$; Table~\ref{tab:main-full}).
H-EPM, meanwhile, outperforms Base in most settings, yet its gain is reliable in only four settings.
Table~\ref{tab:frontier} reveals a similar pattern for proprietary models (GPT-4.1 and GPT-5.1) on ToolSandbox.
Insight, a distinct memory type, is discussed in Section~\ref{sec:insight}.

\begin{figure}[t]
\centering
\includegraphics[width=0.79\columnwidth]{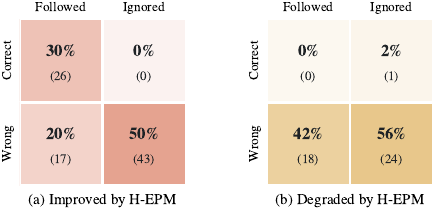}
% \caption{Manual analysis of H-EPM on Mistral-Small-3.2-24B: the decisive turns in the scenarios that H-EPM (a) improves and (b) degrades over Base, by whether the suggestion is correct and followed. Each cell gives the share and number of turns ($n$ = 86 and 43 in total). Figure~\ref{fig:analysis-e4b} shows other models.}
\caption{Decisive turns in the scenarios where H-EPM (a) improves and (b) degrades on Mistral-Small-3.2-24B, by whether the tool suggestion is correct and followed; each cell gives the percentage and number of turns. Figure~\ref{fig:analysis-e4b} shows the results of gemma-4-E4B-it.}
\label{fig:analysis}
\end{figure}
\paragraph{Efficiency.}
As Table~\ref{tab:cost} and Figure~\ref{fig:cost} in the Appendix reveal, H-EPM     on average uses 2.72$\times$ as many LLM calls and 2.21$\times$ as many tokens as Base on ToolSandbox, with a similar increase in LLM calls on BFCL-v4.
ToolNet instead uses fewer tokens than Base, as it presents only a subset of tools to the agent, but is not notably better in any setting.
Neither method therefore offers a better performance--cost trade-off than Base.

\section{Origin of Tool Graph Memory's Gains}
\label{sec:why}
% While tool graph memory offers no consistent improvement, H-EPM does help in a few setups, such as Mistral-Small-3.2-24B on ToolSandbox. 
% We thus examine where these gains truly come from.
% 왜 hepm만 가지고 하는지? 사실 툴넷도 분석해놓긴 했음 어펜딕스에
% 그러나 유의미한 향상이 대부분 hepm에서만 존재했고 그 모델(세팅)들만 여기서 분석함
While tool graph memory offers no consistent improvement, H-EPM does achieve significant gains in a few settings, such as Mistral-Small-3.2-24B on ToolSandbox.
% , whereas ToolNet never does.
We thus focus on these gains and examine where they truly come from.

\subsection{Manual Analysis of Successful Scenarios}
\label{sec:manual}

On Mistral-Small-3.2-24B, we manually inspect the full trajectories of scenarios that fail under Base but succeed with H-EPM. 
We annotate every decisive turn---one that accounts for the outcome difference between H-EPM and Base---on two criteria: (1) the correctness of the suggestion and (2) whether the agent adopted it.

Figure~\ref{fig:analysis}a shows that H-EPM provides a correct tool suggestion in only 30\% of these turns, meaning its recommendations are often misleading.
In another 50\%, the agent calls the necessary tool on its own despite an incorrect suggestion.
The remaining 20\% occur in insufficient-information cases (see Appendix~\ref{app:manual}).
In sum, much of the improvement here does not result from simply following helpful next-tool suggestions.

\subsection{Ablation Study}
\label{sec:control}
We further conduct experiments on three models to test whether H-EPM's suggested tools themselves drive the gain.
We replace them \emph{with random tools} or remove them entirely (\emph{w/o suggestions}), keeping the rest of H-EPM fixed.
% Table~\ref{tab:ablation} shows that both variants perform comparably; the latter even outperforms the original on two of the three models.
Table~\ref{tab:ablation} shows that both variants perform comparably to H-EPM; the latter even outperforms it on two of the three models.

\begin{table}[t]
\centering
\footnotesize
\setlength{\tabcolsep}{2.6pt}
\renewcommand{\arraystretch}{1.15}
\setlength{\aboverulesep}{0pt}
\setlength{\belowrulesep}{0pt}
\resizebox{\columnwidth}{!}{%
\begin{tabular}{l ccc}
\toprule
\textbf{Conditions / Models} & Mistral-Small-3.2-24B & gpt-oss-20b & gemma-4-E4B-it \\
\midrule
Base (only LLMs) & 0.545\hspace{0.7em}\phantom{\fontsize{8}{8}\selectfont(2.0$\times$)} & 0.579\hspace{0.7em}\phantom{\fontsize{8}{8}\selectfont(2.0$\times$)} & 0.588\hspace{0.7em}\phantom{\fontsize{8}{8}\selectfont(2.0$\times$)} \\
Tool graph memory (H-EPM) & 0.603\rlap{$^{+}$}\hspace{0.7em}{\fontsize{8}{8}\selectfont(2.0$\times$)} & 0.620\hspace{0.7em}{\fontsize{8}{8}\selectfont(1.8$\times$)} & 0.554\hspace{0.7em}{\fontsize{8}{8}\selectfont(2.0$\times$)} \\
\rowcolor{gray!15}\multicolumn{4}{l}{\textit{Replacing the suggested tools of H-EPM}\rule[-0.9ex]{0pt}{2.9ex}} \\
\hspace{0.6em}with random tools & 0.571\hspace{0.7em}{\fontsize{8}{8}\selectfont(2.3$\times$)} & 0.663\rlap{$^{+}$}\hspace{0.7em}{\fontsize{8}{8}\selectfont(2.7$\times$)} & 0.553\hspace{0.7em}{\fontsize{8}{8}\selectfont(2.0$\times$)} \\
\hspace{0.6em}with the correct tool (oracle) & 0.682\rlap{$^{+}$}\hspace{0.7em}{\fontsize{8}{8}\selectfont(1.6$\times$)} & 0.643\rlap{$^{+}$}\hspace{0.7em}{\fontsize{8}{8}\selectfont(2.5$\times$)} & 0.666\rlap{$^{+}$}\hspace{0.7em}{\fontsize{8}{8}\selectfont(2.0$\times$)} \\
\rowcolor{gray!15}\multicolumn{4}{l}{\textit{Removing a component of H-EPM}\rule[-0.9ex]{0pt}{2.9ex}} \\
\hspace{0.6em}w/o suggestions & 0.602\rlap{$^{+}$}\hspace{0.7em}{\fontsize{8}{8}\selectfont(1.7$\times$)} & 0.633\rlap{$^{+}$}\hspace{0.7em}{\fontsize{8}{8}\selectfont(2.7$\times$)} & 0.598\hspace{0.7em}{\fontsize{8}{8}\selectfont(1.9$\times$)} \\
\hspace{0.6em}w/o summarization & 0.595\rlap{$^{+}$}\hspace{0.7em}{\fontsize{8}{8}\selectfont(1.8$\times$)} & 0.633\rlap{$^{+}$}\hspace{0.7em}{\fontsize{8}{8}\selectfont(2.6$\times$)} & 0.604\hspace{0.7em}{\fontsize{8}{8}\selectfont(1.8$\times$)} \\
\rowcolor{gray!15}\multicolumn{4}{l}{\textit{Without tool graph memory}\rule[-0.9ex]{0pt}{2.9ex}} \\
\hspace{0.6em}Base + instruction & 0.594\hspace{0.7em}{\fontsize{8}{8}\selectfont(1.9$\times$)} & 0.645\rlap{$^{+}$}\hspace{0.7em}{\fontsize{8}{8}\selectfont(2.7$\times$)} & 0.621\rlap{$^{+}$}\hspace{0.7em}{\fontsize{8}{8}\selectfont(2.0$\times$)} \\
% \hspace{0.6em}Insight memory & 0.620\rlap{$^{+}$}\hspace{0.7em}{\fontsize{8}{8}\selectfont(2.5$\times$)} & 0.630\rlap{$^{+}$}\hspace{0.7em}{\fontsize{8}{8}\selectfont(2.2$\times$)} & 0.620\hspace{0.7em}{\fontsize{8}{8}\selectfont(3.2$\times$)} \\
% \rowcolor{gray!15}\multicolumn{4}{l}{\textit{Adding to insight memory}\rule[-0.9ex]{0pt}{2.9ex}} \\
% \hspace{0.6em}Insight memory + H-EPM & 0.600\rlap{$^{+}$}\hspace{0.7em}{\fontsize{8}{8}\selectfont(2.5$\times$)} & 0.585\hspace{0.7em}{\fontsize{8}{8}\selectfont(2.7$\times$)} & 0.617\hspace{0.7em}{\fontsize{8}{8}\selectfont(2.8$\times$)} \\
% \hspace{0.6em}Insight memory + instruction & 0.671\rlap{$^{+}$}\hspace{0.7em}{\fontsize{8}{8}\selectfont(2.2$\times$)} & 0.675\rlap{$^{+}$}\hspace{0.7em}{\fontsize{8}{8}\selectfont(3.0$\times$)} & 0.640\hspace{0.7em}{\fontsize{8}{8}\selectfont(2.7$\times$)} \\
% \bottomrule
\midrule
\rowcolor{gray!15}\multicolumn{4}{l}{\textit{Adding to insight memory (\S\ref{sec:insightmem})}\rule[-0.9ex]{0pt}{2.9ex}} \\
% \hspace{0.6em}Insight memory & 0.620\rlap{$^{+}$}\hspace{0.7em}{\fontsize{8}{8}\selectfont(2.5$\times$)} & 0.630\rlap{$^{+}$}\hspace{0.7em}{\fontsize{8}{8}\selectfont(2.2$\times$)} & 0.620\hspace{0.7em}{\fontsize{8}{8}\selectfont(3.2$\times$)} \\
\hspace{0.6em}Insight memory + H-EPM & 0.600\rlap{$^{+}$}\hspace{0.7em}{\fontsize{8}{8}\selectfont(2.5$\times$)} & 0.585\hspace{0.7em}{\fontsize{8}{8}\selectfont(2.7$\times$)} & 0.617\hspace{0.7em}{\fontsize{8}{8}\selectfont(2.8$\times$)} \\
\hspace{0.6em}Insight memory + instruction & 0.671\rlap{$^{+}$}\hspace{0.7em}{\fontsize{8}{8}\selectfont(2.2$\times$)} & 0.675\rlap{$^{+}$}\hspace{0.7em}{\fontsize{8}{8}\selectfont(3.0$\times$)} & 0.640\hspace{0.7em}{\fontsize{8}{8}\selectfont(2.7$\times$)} \\
\bottomrule
\end{tabular}}
% \caption{Ablation study of H-EPM on ToolSandbox, with $^{+}$/$^{-}$ as in Table~\ref{tab:main}. Numbers in parentheses denote tool calls relative to Base in the scenarios that each condition flips from failure to success.}
\caption{Ablations on ToolSandbox, with $^{+}$/$^{-}$ as in Table~\ref{tab:main}. Parentheses give the \# of tool calls relative to Base in cases each variant flips from failure to success.}
\label{tab:ablation}
\end{table}

This raises the question of what actually accounts for H-EPM's improvement.
% We find that among tasks that fail under Base, tool calls---invocations of the scenario's tools, excluding H-EPM's summary tool and responses to the user---increase markedly more in those flipped to success than in those still failing, across all conditions.
% Under H-EPM, the former involve about twice as many tool calls as Base (Appendix~\ref{app:flip-trials}).
% We thus hypothesize that the gain mainly arises from the agent simply attempting more tool calls.
For each condition, we take the scenarios that fail under Base and split them into two groups: those the condition turns into successes and those that still fail.
For each group, we count tool calls, i.e., invocations of the available tools, excluding H-EPM's summary tool and responses to the user.
Across all conditions, the flipped scenarios show a much larger increase in tool calls over Base than those that still fail; under H-EPM, they make about twice as many calls as Base (Appendix~\ref{app:flip-trials}).
We thus hypothesize that the gain mainly arises from the agent simply attempting more tool calls.

We test this hypothesis by giving the base agent a single instruction without any tool graph: ``Do not give a final answer or give up until you have tried the available tools'' (\emph{Base + instruction}).
\emph{Base + instruction} attempts more tool calls while adding almost no LLM calls (Table~\ref{tab:flip-trials}) and outperforms H-EPM on two of the three models (Table~\ref{tab:ablation}). 

We conclude that the guidance of tool graph memory contributes little to the performance improvement: a comparable boost can be obtained by encouraging the agent to attempt more tool calls, without the graph and at a much lower inference cost.
% The annotation procedure, an example of an improved scenario, and the control conditions are 
Experimental details are given in Appendix~\ref{app:gains}.

\section{Towards Better Memory Structure}
\label{sec:insight}
\subsection{Lessons from Tool Graph Memory}
\label{sec:modes}
Although the tool suggestions contribute little to the gains of H-EPM (\S\ref{sec:why}), Figure~\ref{fig:analysis} shows that the agent follows every correct suggestion in the improved scenarios, while 98\% of the suggestions at the decisive turns of the degraded scenarios are incorrect.
We thus investigate whether tool graph memory would be more effective if its suggestions were always correct.
To test this, we evaluate an \emph{oracle} condition that replaces H-EPM's prediction with the correct tool at each turn, as an upper bound on tool suggestions.
The \emph{oracle} improves over H-EPM, but many scenarios still fail (Table~\ref{tab:ablation}).

% We classify failures by the decision that caused them, in the order in which the agent makes them (see Appendix~\ref{app:failure}): whether to call a tool at all (\emph{call necessity}), which tool to call (\emph{tool selection}), when to call it (\emph{dependency}), and with what arguments (\emph{argument}). These modes are not mutually exclusive; when a decision fits more than one, we take the earliest in this order. 
% Under the \emph{oracle}, \emph{tool selection} errors almost disappear (4\% of the failures), while the other three modes account for the remaining 96\% (Table~\ref{tab:oracle-fail}). These failures include calling a tool when the agent should respond to or ask the user, or giving up without calling a needed tool (\emph{call necessity}); sending a message before turning on cellular service (\emph{dependency}); and making up dates (\emph{argument}).
We classify each failure by the decision that caused it (see Appendix~\ref{app:failure}): whether to call a tool at all (\emph{call necessity}), which tool to call (\emph{tool selection}), when to call it (\emph{dependency}), and with what arguments (\emph{argument}). These follow the order in which the agent makes decisions, and when a failure fits more than one, we assign the earliest.

Under the \emph{oracle}, where the correct next tool is always suggested, \emph{tool selection} errors almost disappear (4\% of failures), while the other three modes account for the rest (Table~\ref{tab:oracle-fail}).
The remaining failures thus concern what to do in a situation rather than which tool to call next.
A tool graph stores only transitions between tools; encoding such situations would require trajectories that cover them, which is costly in practice.
This motivates memory that provides situation-level guidance rather than direct tool suggestions. 
% A tool graph could in principle be extended to encode such situations individually, but doing so would require far broader trajectory coverage, which is costly and hard to achieve in practice.

\subsection{Insight Memory}
\label{sec:insightmem}
Section~\ref{sec:control} further shows that high-level guidance alone can benefit the agent (\emph{Base + instruction}).
Strategy-based experiential memory~\cite{hu2026memory,jiang2026adaptationagenticaisurvey} distills past trajectories into transferable knowledge, either as workflows~\cite{wang2025agent}, such as tool graphs, or as insights~\cite{zhao2024expel}; the latter captures guidance at the level of situations rather than tools.
We therefore adopt insights as a direct counterpart to tool graph memory, expecting them to address the failure modes that persist even under the \emph{oracle}.

\paragraph{Construction.}
Episodic experience can be consolidated into semantic knowledge in the form of insights~\cite{sumers2024cognitive,liang2025aimeetsbrainmemory,du2026memoryautonomousllmagentsmechanisms}.
% Building on this line of work, 
We construct insight memory (Insight) from H-EPM's episodic summaries, including those of failed trajectories, without using its graph: we cluster the summaries, have the agent model distill what to do and what to avoid for each cluster, and add the insights of the most relevant clusters to the prompt at each turn (Appendix~\ref{app:insight-details}).
The two memories thus draw on the same episodic experience but provide guidance in different forms.
% Because insight memory is built from the same summaries as H-EPM, the two differ only in the form of guidance they provide. 
% Unlike the \emph{oracle}, it also requires no knowledge of the correct tool, so it is usable in practice.

\begin{figure}[t]
\centering
\includegraphics[width=\columnwidth]{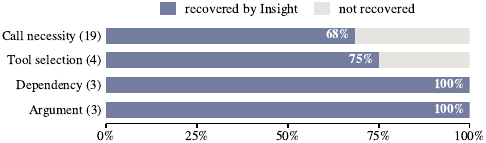}
% \caption{Scenarios that H-EPM degrades on Mistral-Small-3.2-24B ($n$ = 15), by failure mode (\S\ref{sec:modes}), and the fraction recovered by insight memory.  Figure~\ref{fig:analysis-e4b} in the Appendix reports the results of other models.}
% \caption{Scenarios that H-EPM degrades on Mistral-Small-3.2-24B and gpt-oss-20b ($n$ = 29), by failure mode (\S\ref{sec:modes}), and the fraction recovered by Insight.}
\caption{Scenarios that H-EPM degrades on Mistral-Small-3.2-24B and gpt-oss-20b, by failure mode (\S\ref{sec:modes}), and the percentage recovered by insight memory.}
\label{fig:recover}
\end{figure}

\paragraph{Results.}
% Insight achieves the best performance in most settings of Table~\ref{tab:main}, with about twice the average gain of H-EPM over Base and significant gains in more than twice as many settings as both tool graph memories combined; it also achieves the highest pass@3 and pass$^3$ in most settings (Table~\ref{tab:pass}). It keeps LLM calls close to those of Base on both benchmarks and consumes fewer tokens than H-EPM for every model (Table~\ref{tab:cost}). Insight memory also recovers most of the scenarios that H-EPM degrades on Mistral-Small-3.2-24B (Figure~\ref{fig:recover}). In particular, it recovers most \emph{call necessity} failures and every \emph{dependency} and \emph{argument} failure, which next-tool suggestions alone cannot fully address. 
% ExpeL~\cite{zhao2024expel}, which also extracts insights from trajectories, outperforms both tool graph memories (Table~\ref{tab:expel}), suggesting that this advantage is not specific to our insight memory. Even when built only from successful trajectories, as tool graph memory is, insight memory remains competitive with it, while the failed trajectories that tool graph memory discards yield further gains (Table~\ref{tab:succfull}).
Insight performs best in most settings of Table~\ref{tab:main}. Its gains over Base are significant in far more settings than those of either tool graph memory, 
% and it attains the highest pass@3 and pass$^3$ in most cases (Table~\ref{tab:pass}).
and it attains the highest pass@3 in almost all settings, with competitive pass$^3$ (Table~\ref{tab:pass}).
It keeps LLM calls close to Base and uses fewer tokens than H-EPM across models (Table~\ref{tab:cost}). It also recovers most scenarios that H-EPM degrades on Mistral-Small-3.2-24B and gpt-oss-20b, including every \emph{dependency} and \emph{argument} failure (Figure~\ref{fig:recover}). This advantage is not implementation-specific: ExpeL~\cite{zhao2024expel} also beats both tool graph memories (Table~\ref{tab:expel}), and insights built only from successful trajectories, as tool graph memory is, achieve comparable performance (Table~\ref{tab:succfull}).

\paragraph{Insight + tool graph.}
Contrary to expectations, \emph{Insight memory + H-EPM} performs worse than insight memory alone on seven of the nine models and requires 2.68$\times$ as many LLM calls (Table~\ref{tab:comboruns}). In most scenarios that insight memory solves but the combination fails, the failure patterns of tool graph memory recur, and none is caused by insight alone. \emph{Insight memory + instruction} performs best among all conditions without the correct next tool on all three models, even surpassing \emph{oracle} on gpt-oss-20b (Table~\ref{tab:ablation}), as the instruction discourages giving up, the most common failure under \emph{oracle}. 
In summary, these results support insight memory as a practical alternative to tool graph memory for multi-turn tool use. It distills situation-level guidance from the same experience and outperforms tool graph memory in most of our settings. Examples and further analyses are in Appendix~\ref{app:failure}.
% These results suggest that what the agent needs may be higher-level guidance on its decisions rather than next-tool suggestions. 
% Taken together, these results favor situation-level gui9dance over next-tool suggestions as the form of memory for multi-turn tool use. Examples and further analyses are given in Appendix~\ref{app:failure}.

\section{Conclusion}
% We systemically evaluate tool graph memory on nine models and two multi-turn tool-use benchmarks. 
% We demonstrate that it does not consistently improve over baselines, and the gains of the best method, i.e., H-EPM, come at more than twice as many LLM calls. 
% These gains are associated with attempting more tool calls rather than with following correct suggestions, and a simple instruction yields comparable gains. 
% Even with the correct next tool, many failures remain, such as giving up without calling a necessary tool.
% Insight memory, which provides higher-level guidance on what to do in the current situation, outperforms tool graph memory in most settings with almost no additional LLM calls.
% Memory for multi-turn tool use may thus be more useful when it provides such guidance than when it suggests the next tool.
Our results raise the question of what guidance agents need from memory in multi-turn tool use. 
Tool graph memory suggests the next tool from past transitions, yet random or no suggestions retain most of its gains, and even correct suggestions leave many failures: tool transitions alone do not tell agents whether, when, or with what arguments to call tools. 
Insight memory instead distills the same experience into semantic knowledge about what to do in a situation, which transfers across scenarios, and outperforms tool graph memory at almost no additional LLM calls. 
These outcomes suggest that semantic memory offers a better performance--cost trade-off than procedural or episodic memory, by informing the agent's decisions rather than prescribing its next tool.

\section*{Limitations}
% Our study focuses on tool graph memory, in which nodes are tools and edges represent transitions between them, as in ToolNet and H-EPM. We therefore make no claim about graph memories whose nodes represent entities other than tools, such as states or arguments. 

% We use a simple implementation of insight memory, as our goal is to test whether higher-level guidance helps rather than to optimize the memory itself. Accordingly, we do not systematically study its sensitivity to design choices such as the encoder and the clustering method, and we select the number of clusters $K$ by the silhouette score rather than by task performance. Likewise, our cost comparison covers inference only and does not include the offline cost of constructing the memories. For insight memory, this construction reuses H-EPM's episodic summaries and adds clustering and LLM-based insight extraction.

% Finally, our evaluation covers two multi-turn tool-use benchmarks, nine open-weight models from 2B to 120B parameters, and two frontier models on ToolSandbox. Evaluation on additional benchmarks, other tool environments, and more frontier models would further test the generality of our findings.
Our analysis targets tool graph memory, in which nodes are tools and edges are transitions between them, as in ToolNet and H-EPM. Tool graph memory is one form of graph memory; graph memories whose nodes and edges also represent states, arguments, or other entities go beyond what we target here. What our analysis offers to such memories is a lesson: insights about what to do in a situation can help agents more than next-tool suggestions, and our failure-mode analysis suggests what richer graph memories may need to capture.

Insight memory is kept deliberately lightweight, since our aim is to test whether situation-level guidance helps relative to tool graph memory rather than to optimize the memory itself: we build it without training, using a pretrained sentence encoder and standard $K$-means, with $K$ selected by the silhouette score rather than by task performance, and its construction adds clustering and LLM-based insight extraction on top of H-EPM's episodic summaries, an offline cost incurred once. That insight memory outperforms tool graph memory in most settings even in this lightweight form supports the value of situation-level guidance, and leaves richer experience and other extraction strategies for future work.

Our evaluation spans two multi-turn tool-use benchmarks, nine open-weight models from 2B to 120B parameters, and two frontier models on ToolSandbox, with repeated runs and paired tests. The analyses that explain H-EPM's gains are narrower: the ablations and manual analyses focus on the same three models on ToolSandbox, chosen to cover two of its larger gains and one decrease. The qualitative pattern is consistent across the three.

{\catcode`\&=12\relax
\bibliography{custom}}
\clearpage
\appendix
\raggedbottom
\setcounter{topnumber}{3}\setcounter{dbltopnumber}{3}\renewcommand{\topfraction}{0.95}\renewcommand{\dbltopfraction}{0.95}\renewcommand{\bottomfraction}{0.6}\renewcommand{\textfraction}{0.05}\renewcommand{\floatpagefraction}{0.5}\renewcommand{\dblfloatpagefraction}{0.5}

\FloatBarrier
\section{Detailed Experimental Setup}
\label{app:setup}

\subsection{Benchmarks and Evaluation}
ToolSandbox~\cite{lu-etal-2025-toolsandbox} is a stateful, conversational benchmark with 34 tools over 11 domains. Its 1{,}032 scenarios include distraction tools and test state dependencies, canonicalization, and insufficient information, in which a needed tool is withheld and the agent should recognize that it cannot complete the task. We use the training/test split of H-EPM (825 training and 207 test scenarios).
A trajectory is scored by its similarity to the milestones, events that must occur, and scores zero if it matches a minefield, an event that must not occur; following H-EPM, a scenario counts as a success when this score exceeds 0.55. The multi-turn subset of BFCL-v4~\cite{patil2025the} has 800 tasks over eight APIs in four categories: base, missing function, missing parameter, and long context. We hold out all four variants of 40 tasks as the test set (160 tasks) and use the other 640 for training, so that no underlying task is shared. A task is correct only if the backend state and the response checks pass at every turn.

Following ToolSandbox, we use GPT-4o~\cite{openai2024gpt4ocard} as the user simulator. With GPT-4.1 instead, as in the H-EPM protocol, for both memory construction and evaluation on Qwen3-8B and gemma-4-E4B-it, neither ToolNet nor H-EPM is significantly above Base, and Insight scores highest on both (Table~\ref{tab:usersim}).

\begin{table}[H]
\centering
\footnotesize
\renewcommand{\arraystretch}{1.15}
\setlength{\tabcolsep}{3.2pt}
\setlength{\aboverulesep}{0pt}
\setlength{\belowrulesep}{0pt}
\resizebox{\columnwidth}{!}{%
\begin{tabular}{l ccc ccc}
\toprule
 & \multicolumn{3}{c}{Qwen3-8B} & \multicolumn{3}{c}{gemma-4-E4B-it} \\
\cmidrule(lr){2-4}\cmidrule(lr){5-7}
\textbf{Method} & Score & $\Delta$ & $p$ & Score & $\Delta$ & $p$ \\
\midrule
Base & 0.462 & -- & -- & 0.575 & -- & -- \\
ToolNet & 0.483 & +0.021 & 0.364 & 0.573 & $-$0.002 & 0.920 \\
H-EPM & 0.496 & +0.034 & 0.277 & 0.559 & $-$0.016 & 0.573 \\
Insight & \textbf{0.527} & +0.065 & 0.023 & \textbf{0.583} & +0.009 & 0.740 \\
\bottomrule
\end{tabular}}
\caption{ToolSandbox with GPT-4.1 as the user simulator for both memory construction and evaluation ($n{=}207$ scenarios). $\Delta$ is the difference from Base, computed before rounding; significance is tested as in Table~\ref{tab:main-full}.}
\label{tab:usersim}
\end{table}

\subsection{Implementation Details}
H-EPM also gives the agent a summarization tool, \texttt{summarize\_the\_task}, with a system-prompt line asking it to call the tool; each use of the tool adds an LLM call, and its summary accumulates in the context. All of H-EPM's additional calls and the insight extraction use the agent model itself. To build the memories, we roll out each training scenario three times and keep the highest-scoring rollout, or the shortest one on ties.

The agent models are Qwen3.5-2B~\cite{qwen35blog}, Qwen3-8B~\cite{yang2025qwen3technicalreport}, Qwen3.6-35B-A3B~\cite{qwen36_35b_a3b}, gemma-4-E4B-it and gemma-4-26B-A4B-it~\cite{gemmateam2026gemma4technicalreport}, Ministral-3-8B~\cite{liu2026ministral3}, Mistral-Small-3.2-24B~\cite{ref-mistralsmall32}, and gpt-oss-20b and gpt-oss-120b~\cite{openai2025gptoss120bgptoss20bmodel}; the frontier models in Table~\ref{tab:frontier} are GPT-4.1~\cite{openai2025gpt41} and GPT-5.1~\cite{openai2025gpt51}. We use greedy decoding (temperature 0) for both the agent and the user simulator and serve all open-weight models with vLLM~\cite{kwon2023efficientmemorymanagementlarge} on a single NVIDIA H200 GPU.

\subsection{Insight Memory Details}
\label{app:insight-details}
Insight memory is built from the episodic summaries of H-EPM, including those of failed trajectories; H-EPM writes a summary only when an LLM call judges it necessary, so summaries exist for 7\% to 26\% of the 825 ToolSandbox training scenarios, depending on the model. Using failed trajectories is common in insight memory: ExpeL~\cite{zhao2024expel} and ReasoningBank~\cite{ouyang2026reasoningbank} learn from both failed and successful trajectories, and Reflexion~\cite{shinn2023reflexion} from failed attempts, whereas tool graph memory uses successful trajectories only. 

We embed the summaries with all-MiniLM-L6-v2~\cite{wang2020minilmdeepselfattentiondistillation} and cluster them with $K$-means, choosing $K$ among 15 candidates from 5 to 150 by the silhouette score, subject to clusters holding more than three summaries on average (Table~\ref{tab:kselect}). The agent model then extracts insights from each cluster with the prompt in Figure~\ref{fig:prompt}.

At every model call, the five most recent messages are embedded with the same encoder and compared with all summaries, and the insights of the clusters of the two most similar summaries are added to the system prompt.
\begin{table}[!htb]
\centering
\footnotesize
\setlength{\tabcolsep}{5pt}
\setlength{\aboverulesep}{0pt}
\setlength{\belowrulesep}{0pt}
\renewcommand{\arraystretch}{1.15}
\begin{tabular}{lrrr}
\toprule
\textbf{Model} & $n$ & $K$ & $n/K$ \\
\midrule
\rowcolor{gray!15}\rlap{\textit{ToolSandbox}}\rule[-0.9ex]{0pt}{2.9ex}&&& \\
Qwen3.5-2B & 77 & 25 & 3.1 \\
Qwen3-8B & 360 & 100 & 3.6 \\
Qwen3.6-35B-A3B & 146 & 5 & 29.2 \\
gemma-4-E4B-it & 261 & 15 & 17.4 \\
gemma-4-26B-A4B-it & 181 & 5 & 36.2 \\
Ministral-3-8B & 300 & 5 & 60.0 \\
Mistral-Small-3.2-24B & 109 & 8 & 13.6 \\
gpt-oss-20b & 170 & 5 & 34.0 \\
gpt-oss-120b & 185 & 15 & 12.3 \\
\midrule
\rowcolor{gray!15}\rlap{\textit{BFCL-v4}}\rule[-0.9ex]{0pt}{2.9ex}&&& \\
Qwen3.5-2B & 26,877 & 30 & 895.9 \\
Qwen3-8B & 2,235 & 150 & 14.9 \\
Qwen3.6-35B-A3B & 46,738 & 40 & 1168.5 \\
gemma-4-E4B-it & 489 & 80 & 6.1 \\
gemma-4-26B-A4B-it & 924 & 50 & 18.5 \\
Ministral-3-8B & 5,247 & 100 & 52.5 \\
Mistral-Small-3.2-24B & 2,253 & 150 & 15.0 \\
gpt-oss-20b & 122 & 40 & 3.1 \\
gpt-oss-120b & 787 & 5 & 157.4 \\
\bottomrule
\end{tabular}
\caption{Number of clusters selected per model. $n$ is the number of episodic summaries clustered, $K$ the selected number of clusters, and $n/K$ the mean number of summaries per cluster.}
\label{tab:kselect}
\end{table}

\begin{figure}[H]
\centering
\begin{tcolorbox}[title=\textbf{Insight Extraction Prompt},
colback=gray!5,colframe=gray!40,coltitle=black,colbacktitle=gray!18,
fonttitle=\small,boxrule=0.4pt,left=4pt,right=4pt,top=4pt,bottom=4pt]
\ttfamily\fontsize{6.5}{7.5}\selectfont\setlength{\parskip}{2pt}\raggedright
You are mining INSIGHTS from a cluster of AI tool-use trajectories. Each trajectory carries a
TRAJECTORY-LEVEL tag [SUCCESS] or [FAILURE] --- it reflects ONLY whether the trajectory's final
state matched the user's intent. It says nothing about individual moves.\\[4pt]
Extract concrete, transferable insights for similar future situations:\\
\hspace*{1em}- "do": a move that demonstrably WORKED. Mine from repeated successful patterns AND
from correct early moves of failed trajectories.\\
\hspace*{1em}- "avoid": a move that demonstrably HURT. Mine from concrete failure patterns AND from
wasted detours inside successes.\\[4pt]
Every insight MUST cite a specific observable signal from the summaries: tool name + argument shape,
or environment state, or user response pattern, or sub-sequence of 2+ tool calls. The
SUCCESS/FAILURE tag is NEVER a signal.\\[4pt]
Support thresholds: DO needs $\ge$2 supporting trajectories. AVOID needs $\ge$1 trajectory IF the
failure mode names a concrete artifact.\\[4pt]
If only 1 trajectory or only 1 distinct pattern is present, emit FEWER insights. An empty list is
acceptable. Do NOT pad to a count. Hard cap 10. Single actionable sentence per insight, max
$\sim$30 words.\\[4pt]
\{trajectories\_block\}\\[4pt]
Return JSON only.
\end{tcolorbox}
\caption{The prompt used to extract insights from each cluster.}
\label{fig:prompt}
\end{figure}

\FloatBarrier
\section{Full Results}
\label{app:full}
\begin{figure}[!htb]
\centering
\includegraphics[width=0.85\columnwidth]{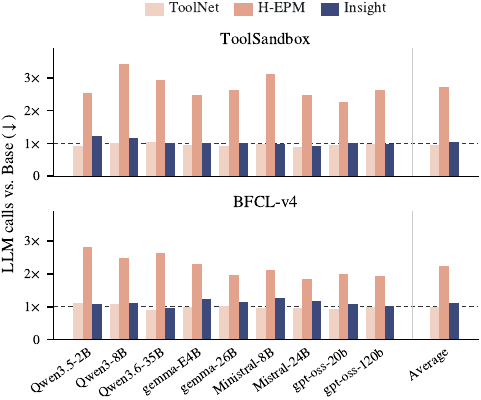}
\caption{LLM calls per scenario relative to Base of the same model. The Average bars give the ratios of the nine-model averages.}
\label{fig:cost}
\end{figure}

Tables~\ref{tab:frontier}--\ref{tab:cost} and Figure~\ref{fig:cost} report the full results: frontier models, per-model scores with significance tests, pass rates, and inference cost.
\begin{table}[H]
\centering
\footnotesize
\renewcommand{\arraystretch}{1.2}
\setlength{\tabcolsep}{6pt}
\begin{tabular}{l cccc}
\toprule
\textbf{Model} & \textbf{Base} & \textbf{ToolNet} & \textbf{H-EPM} & \textbf{Insight} \\
\midrule
GPT-4.1 & 0.724 & 0.709 & 0.725 & \textbf{0.729} \\
GPT-5.1 & 0.701 & 0.694 & \textbf{0.702} & \textbf{0.702} \\
\bottomrule
\end{tabular}
\caption{Task performance of frontier models on ToolSandbox ($n$ = 207).}
\label{tab:frontier}
\end{table}

\begin{table*}[!tbp]
\centering
\footnotesize
\setlength{\tabcolsep}{3pt}
\renewcommand{\arraystretch}{1.15}
\setlength{\aboverulesep}{0pt}
\setlength{\belowrulesep}{0pt}
\resizebox{\textwidth}{!}{%
\begin{tabular}{l c c r@{\hspace{3pt}}l c r@{\hspace{3pt}}l c r@{\hspace{3pt}}l}
\toprule
\textbf{Model} & \textbf{Base} & \textbf{ToolNet} & \multicolumn{2}{c}{$\Delta$ ($p$-value)} & \textbf{H-EPM} & \multicolumn{2}{c}{$\Delta$ ($p$-value)} & \textbf{Insight} & \multicolumn{2}{c}{$\Delta$ ($p$-value)} \\
\midrule
\rowcolor{gray!15}\rlap{\textit{ToolSandbox}}\rule[-0.9ex]{0pt}{2.9ex}&&&&&&&&&& \\
Qwen3.5-2B & 0.420$_{\pm 0.003}$ & 0.428$_{\pm 0.008}$ & $+$0.008 & {\scriptsize (0.399)} & 0.436$_{\pm 0.015}$ & $+$0.016 & {\scriptsize (0.469)} & \textbf{0.488}$_{\pm 0.009}$ & $+$0.068 & {\scriptsize (0.015)} \\
Qwen3-8B & 0.500$_{\pm 0.008}$ & 0.507$_{\pm 0.009}$ & $+$0.006 & {\scriptsize (0.660)} & 0.476$_{\pm 0.010}$ & $-$0.024 & {\scriptsize (0.405)} & \textbf{0.556}$_{\pm 0.004}$ & $+$0.056 & {\scriptsize (0.037)} \\
Qwen3.6-35B-A3B & 0.626$_{\pm 0.001}$ & 0.608$_{\pm 0.002}$ & $-$0.018 & {\scriptsize (0.256)} & 0.670$_{\pm 0.005}$ & $+$0.044 & {\scriptsize (0.055)} & \textbf{0.674}$_{\pm 0.006}$ & $+$0.048 & {\scriptsize (0.056)} \\
\addlinespace[0.7pt]
gemma-4-E4B-it & 0.588$_{\pm 0.012}$ & 0.547$_{\pm 0.008}$ & $-$0.041 & {\scriptsize (0.004)} & 0.554$_{\pm 0.012}$ & $-$0.034 & {\scriptsize (0.209)} & \textbf{0.620}$_{\pm 0.003}$ & $+$0.032 & {\scriptsize (0.267)} \\
gemma-4-26B-A4B-it & 0.690$_{\pm 0.000}$ & 0.655$_{\pm 0.003}$ & $-$0.035 & {\scriptsize (0.026)} & \textbf{0.705}$_{\pm 0.000}$ & $+$0.016 & {\scriptsize (0.394)} & \textbf{0.705}$_{\pm 0.009}$ & $+$0.015 & {\scriptsize (0.423)} \\
\addlinespace[0.7pt]
Ministral-3-8B & 0.631$_{\pm 0.008}$ & 0.588$_{\pm 0.005}$ & $-$0.044 & {\scriptsize (0.004)} & \textbf{0.672}$_{\pm 0.011}$ & $+$0.041 & {\scriptsize (0.024)} & 0.539$_{\pm 0.009}$ & $-$0.092 & {\scriptsize ($<$0.001)} \\
Mistral-Small-3.2-24B & 0.545$_{\pm 0.003}$ & 0.557$_{\pm 0.019}$ & $+$0.012 & {\scriptsize (0.479)} & 0.603$_{\pm 0.009}$ & $+$0.058 & {\scriptsize (0.014)} & \textbf{0.620}$_{\pm 0.021}$ & $+$0.075 & {\scriptsize (0.004)} \\
\addlinespace[0.7pt]
gpt-oss-20b & 0.579$_{\pm 0.007}$ & 0.573$_{\pm 0.010}$ & $-$0.007 & {\scriptsize (0.721)} & 0.620$_{\pm 0.035}$ & $+$0.041 & {\scriptsize (0.087)} & \textbf{0.630}$_{\pm 0.008}$ & $+$0.050 & {\scriptsize (0.046)} \\
gpt-oss-120b & 0.726$_{\pm 0.012}$ & 0.674$_{\pm 0.016}$ & $-$0.052 & {\scriptsize (0.006)} & 0.713$_{\pm 0.006}$ & $-$0.013 & {\scriptsize (0.483)} & \textbf{0.734}$_{\pm 0.008}$ & $+$0.009 & {\scriptsize (0.655)} \\
\cmidrule(lr){1-11}
\textit{Average} & -- & -- & $-$0.019 &  & -- & $+$0.016 &  & -- & $+$0.029 &  \\
\midrule
\rowcolor{gray!15}\rlap{\textit{BFCL-v4}}\rule[-0.9ex]{0pt}{2.9ex}&&&&&&&&&& \\
Qwen3.5-2B & 0.133$_{\pm 0.016}$ & 0.044$_{\pm 0.000}$ & $-$0.090 & {\scriptsize ($<$0.001)} & 0.138$_{\pm 0.006}$ & $+$0.004 & {\scriptsize (0.803)} & \textbf{0.165}$_{\pm 0.010}$ & $+$0.031 & {\scriptsize (0.166)} \\
Qwen3-8B & 0.171$_{\pm 0.013}$ & 0.129$_{\pm 0.007}$ & $-$0.042 & {\scriptsize (0.156)} & 0.179$_{\pm 0.010}$ & $+$0.008 & {\scriptsize (0.573)} & \textbf{0.242}$_{\pm 0.007}$ & $+$0.071 & {\scriptsize (0.007)} \\
Qwen3.6-35B-A3B & 0.181$_{\pm 0.006}$ & 0.108$_{\pm 0.007}$ & $-$0.073 & {\scriptsize (0.006)} & 0.156$_{\pm 0.006}$ & $-$0.025 & {\scriptsize (0.140)} & \textbf{0.206}$_{\pm 0.000}$ & $+$0.025 & {\scriptsize (0.241)} \\
\addlinespace[0.7pt]
gemma-4-E4B-it & 0.221$_{\pm 0.013}$ & 0.094$_{\pm 0.000}$ & $-$0.127 & {\scriptsize ($<$0.001)} & 0.252$_{\pm 0.016}$ & $+$0.031 & {\scriptsize (0.144)} & \textbf{0.285}$_{\pm 0.004}$ & $+$0.065 & {\scriptsize (0.006)} \\
gemma-4-26B-A4B-it & \textbf{0.566}$_{\pm 0.004}$ & 0.369$_{\pm 0.000}$ & $-$0.197 & {\scriptsize ($<$0.001)} & 0.519$_{\pm 0.000}$ & $-$0.047 & {\scriptsize (0.059)} & 0.531$_{\pm 0.018}$ & $-$0.034 & {\scriptsize (0.265)} \\
\addlinespace[0.7pt]
Ministral-3-8B & 0.209$_{\pm 0.004}$ & 0.141$_{\pm 0.004}$ & $-$0.069 & {\scriptsize (0.026)} & \textbf{0.328}$_{\pm 0.013}$ & $+$0.119 & {\scriptsize ($<$0.001)} & 0.306$_{\pm 0.006}$ & $+$0.097 & {\scriptsize (0.003)} \\
Mistral-Small-3.2-24B & 0.231$_{\pm 0.006}$ & 0.127$_{\pm 0.013}$ & $-$0.104 & {\scriptsize ($<$0.001)} & 0.250$_{\pm 0.011}$ & $+$0.019 & {\scriptsize (0.352)} & \textbf{0.338}$_{\pm 0.019}$ & $+$0.106 & {\scriptsize ($<$0.001)} \\
\addlinespace[0.7pt]
gpt-oss-20b & 0.325$_{\pm 0.013}$ & 0.152$_{\pm 0.019}$ & $-$0.173 & {\scriptsize ($<$0.001)} & 0.369$_{\pm 0.019}$ & $+$0.044 & {\scriptsize (0.100)} & \textbf{0.427}$_{\pm 0.019}$ & $+$0.102 & {\scriptsize ($<$0.001)} \\
gpt-oss-120b & 0.494$_{\pm 0.018}$ & 0.275$_{\pm 0.009}$ & $-$0.219 & {\scriptsize ($<$0.001)} & \textbf{0.559}$_{\pm 0.022}$ & $+$0.066 & {\scriptsize (0.016)} & 0.454$_{\pm 0.024}$ & $-$0.040 & {\scriptsize (0.263)} \\
\cmidrule(lr){1-11}
\textit{Average} & -- & -- & $-$0.121 &  & -- & $+$0.024 &  & -- & $+$0.047 &  \\
\bottomrule
\end{tabular}}
\caption{Full version of Table~\ref{tab:main} ($n$ = 207 scenarios and 160 tasks), with the standard deviation over the three runs in subscript. $\Delta$ is the difference from Base, computed before rounding, with the $p$-value of a two-sided paired $t$-test over scenarios (ToolSandbox) or tasks (BFCL-v4) in parentheses. The Average row gives the mean $\Delta$ over the nine models. The best method per benchmark is in \textbf{bold}.}
\label{tab:main-full}
\end{table*}

\begin{table*}[!tbp]
\centering
\footnotesize
\setlength{\tabcolsep}{5pt}
\renewcommand{\arraystretch}{1.15}
\setlength{\aboverulesep}{0pt}
\setlength{\belowrulesep}{0pt}
\begin{tabular}{l cc cc cc cc}
\toprule
\multicolumn{1}{c}{\multirow{2}{*}{\textbf{Model}}} & \multicolumn{2}{c}{\textbf{Base}} & \multicolumn{2}{c}{\textbf{ToolNet}} & \multicolumn{2}{c}{\textbf{H-EPM}} & \multicolumn{2}{c}{\textbf{Insight}} \\
\cmidrule(lr){2-3}\cmidrule(lr){4-5}\cmidrule(lr){6-7}\cmidrule(lr){8-9}
 & \textbf{pass@3} & \textbf{pass$^3$} & \textbf{pass@3} & \textbf{pass$^3$} & \textbf{pass@3} & \textbf{pass$^3$} & \textbf{pass@3} & \textbf{pass$^3$} \\
\midrule
\rowcolor{gray!15}\rlap{\textit{ToolSandbox}}\rule[-0.9ex]{0pt}{2.9ex}&&&&&&&& \\
Qwen3.5-2B & 0.401 & 0.362 & 0.406 & 0.372 & 0.411 & 0.348 & \textbf{0.488} & \textbf{0.415} \\
Qwen3-8B & 0.546 & 0.425 & 0.527 & 0.435 & 0.527 & 0.449 & \textbf{0.594} & \textbf{0.493} \\
Qwen3.6-35B-A3B & 0.667 & 0.589 & 0.618 & 0.556 & 0.715 & 0.618 & \textbf{0.729} & \textbf{0.628} \\
\addlinespace[0.7pt]
gemma-4-E4B-it & 0.628 & 0.488 & 0.546 & 0.469 & 0.580 & 0.527 & \textbf{0.676} & \textbf{0.556} \\
gemma-4-26B-A4B-it & 0.758 & 0.705 & 0.667 & 0.647 & \textbf{0.783} & \textbf{0.725} & \textbf{0.783} & 0.710 \\
\addlinespace[0.7pt]
Ministral-3-8B & 0.720 & 0.502 & 0.618 & 0.444 & \textbf{0.783} & \textbf{0.647} & 0.638 & 0.406 \\
Mistral-Small-3.2-24B & 0.638 & 0.382 & 0.604 & 0.459 & 0.715 & 0.502 & \textbf{0.739} & \textbf{0.507} \\
\addlinespace[0.7pt]
gpt-oss-20b & 0.667 & 0.473 & 0.633 & 0.454 & 0.739 & 0.541 & \textbf{0.758} & \textbf{0.556} \\
gpt-oss-120b & 0.807 & \textbf{0.705} & 0.725 & 0.618 & \textbf{0.836} & 0.686 & \textbf{0.836} & \textbf{0.705} \\
\midrule
\rowcolor{gray!15}\rlap{\textit{BFCL-v4}}\rule[-0.9ex]{0pt}{2.9ex}&&&&&&&& \\
% Qwen3.5-2B & 0.156 & 0.094 & 0.050 & 0.038 & 0.175 & 0.094 & \textbf{0.238} & \textbf{0.100} \\
% Qwen3-8B & 0.200 & 0.131 & 0.125 & 0.125 & 0.238 & 0.113 & \textbf{0.325} & \textbf{0.163} \\
% Qwen3.6-35B-A3B & 0.206 & 0.150 & 0.138 & 0.075 & 0.181 & 0.125 & \textbf{0.256} & \textbf{0.156} \\
% \addlinespace[0.7pt]
% gemma-4-E4B-it & 0.225 & 0.206 & 0.094 & 0.094 & 0.269 & \textbf{0.219} & \textbf{0.381} & 0.194 \\
% gemma-4-26B-A4B-it & 0.581 & \textbf{0.550} & 0.369 & 0.369 & 0.544 & 0.494 & \textbf{0.594} & 0.469 \\
% \addlinespace[0.7pt]
% Ministral-3-8B & 0.269 & 0.150 & 0.169 & 0.113 & \textbf{0.381} & \textbf{0.275} & \textbf{0.381} & 0.238 \\
% Mistral-Small-3.2-24B & 0.263 & 0.194 & 0.150 & 0.094 & 0.281 & 0.206 & \textbf{0.413} & \textbf{0.281} \\
% \addlinespace[0.7pt]
% gpt-oss-20b & 0.425 & 0.213 & 0.175 & 0.113 & 0.481 & 0.238 & \textbf{0.506} & \textbf{0.369} \\
% gpt-oss-120b & 0.556 & 0.431 & 0.356 & 0.194 & 0.619 & \textbf{0.500} & \textbf{0.638} & 0.288 \\
Qwen3.5-2B & 0.175 & \textbf{0.094} & 0.050 & 0.038 & 0.194 & \textbf{0.094} & \textbf{0.281} & 0.069 \\
Qwen3-8B & 0.238 & 0.113 & 0.150 & 0.113 & 0.256 & 0.113 & \textbf{0.338} & \textbf{0.150} \\
Qwen3.6-35B-A3B & 0.225 & \textbf{0.150} & 0.138 & 0.075 & 0.200 & 0.125 & \textbf{0.288} & 0.125 \\
\addlinespace[0.7pt]
gemma-4-E4B-it & 0.244 & 0.200 & 0.094 & 0.094 & 0.294 & \textbf{0.219} & \textbf{0.381} & 0.181 \\
gemma-4-26B-A4B-it & 0.581 & \textbf{0.550} & 0.369 & 0.369 & 0.544 & 0.494 & \textbf{0.594} & 0.469 \\
\addlinespace[0.7pt]
Ministral-3-8B & 0.269 & 0.150 & 0.169 & 0.113 & 0.381 & \textbf{0.275} & \textbf{0.413} & 0.200 \\
Mistral-Small-3.2-24B & 0.288 & 0.175 & 0.169 & 0.088 & 0.300 & 0.194 & \textbf{0.456} & \textbf{0.244} \\
\addlinespace[0.7pt]
gpt-oss-20b & 0.475 & 0.169 & 0.219 & 0.100 & 0.519 & 0.206 & \textbf{0.563} & \textbf{0.288} \\
gpt-oss-120b & 0.556 & 0.431 & 0.356 & 0.194 & 0.619 & \textbf{0.500} & \textbf{0.694} & 0.231 \\
\bottomrule
\end{tabular}
\caption{pass@3 (solved in at least one of three runs) and pass$^3$ (solved in all three runs) on ToolSandbox ($n$ = 207) and BFCL-v4 ($n$ = 160). Best per model and per metric in \textbf{bold}.}
\label{tab:pass}
\end{table*}

\paragraph{Retrieval latency.}
Insight retrieval requires no additional LLM call and uses the same all-MiniLM-L6-v2 encoder as H-EPM's episodic matching. We measure retrieval latency in-process, excluding model loading, over all agent turns in the 207 test scenarios for gpt-oss-20b, Mistral-Small-3.2-24B, and gemma-4-E4B-it (1{,}295--1{,}494 turns per model). On an AMD EPYC 9655 host, with four CPU threads per shard and 12 concurrent shards sharing the host, the full retrieval path---query embedding followed by similarity scoring and top-2 selection---has a median latency of 15--17 ms per turn (p95 29--39 ms). Embedding accounts for nearly all of this latency, with a median of 15--17 ms (p95 28--37 ms), compared with median LLM-call latencies of 0.4--1.1 s in the same runs. These measurements isolate retrieval overhead rather than end-to-end latency, which also includes prompt prefill and generation. 

% \begin{table*}[!tbp]
\begin{table*}[!t]
\centering
\fontsize{8}{9.6}\selectfont
\setlength{\tabcolsep}{5pt}
\renewcommand{\arraystretch}{1.15}
\setlength{\aboverulesep}{0pt}
\setlength{\belowrulesep}{0pt}
\begin{tabular}{l cccc}
\toprule
\textbf{Model} & \textbf{Base} & \textbf{ToolNet} & \textbf{H-EPM} & \textbf{Insight} \\
\midrule
\rowcolor{gray!15}\rlap{\textit{ToolSandbox LLM calls}}\rule[-0.9ex]{0pt}{2.9ex}&&&& \\
Qwen3.5-2B & 5.42 & 5.04 {\scriptsize (0.93)} & 13.80 {\scriptsize (2.55)} & 6.66 {\scriptsize (1.23)} \\
Qwen3-8B & 4.78 & 4.79 {\scriptsize (1.00)} & 16.36 {\scriptsize (3.42)} & 5.56 {\scriptsize (1.16)} \\
Qwen3.6-35B-A3B & 5.75 & 6.03 {\scriptsize (1.05)} & 16.92 {\scriptsize (2.94)} & 5.74 {\scriptsize (1.00)} \\
\addlinespace[0.7pt]
gemma-4-E4B-it & 5.27 & 4.97 {\scriptsize (0.94)} & 13.00 {\scriptsize (2.47)} & 5.26 {\scriptsize (1.00)} \\
gemma-4-26B-A4B-it & 5.90 & 5.52 {\scriptsize (0.93)} & 15.47 {\scriptsize (2.62)} & 6.04 {\scriptsize (1.02)} \\
\addlinespace[0.7pt]
Ministral-3-8B & 5.71 & 5.51 {\scriptsize (0.97)} & 17.85 {\scriptsize (3.13)} & 5.64 {\scriptsize (0.99)} \\
Mistral-Small-3.2-24B & 5.32 & 4.70 {\scriptsize (0.88)} & 13.24 {\scriptsize (2.49)} & 4.88 {\scriptsize (0.92)} \\
\addlinespace[0.7pt]
gpt-oss-20b & 4.91 & 4.64 {\scriptsize (0.94)} & 11.16 {\scriptsize (2.27)} & 4.99 {\scriptsize (1.02)} \\
gpt-oss-120b & 5.26 & 5.15 {\scriptsize (0.98)} & 13.82 {\scriptsize (2.63)} & 5.10 {\scriptsize (0.97)} \\
\cmidrule(lr){1-5}
\textit{Average} & 5.37 & 5.15 {\scriptsize (0.96)} & 14.62 {\scriptsize (2.72)} & 5.54 {\scriptsize (1.03)} \\
\midrule
\rowcolor{gray!15}\rlap{\textit{ToolSandbox tokens}}\rule[-0.9ex]{0pt}{2.9ex}&&&& \\
Qwen3.5-2B & 10,057 & 7,009 {\scriptsize (0.70)} & 16,965 {\scriptsize (1.69)} & 16,298 {\scriptsize (1.62)} \\
Qwen3-8B & 7,556 & 5,425 {\scriptsize (0.72)} & 25,853 {\scriptsize (3.42)} & 11,324 {\scriptsize (1.50)} \\
Qwen3.6-35B-A3B & 11,289 & 9,010 {\scriptsize (0.80)} & 25,517 {\scriptsize (2.26)} & 15,094 {\scriptsize (1.34)} \\
\addlinespace[0.7pt]
gemma-4-E4B-it & 7,821 & 5,291 {\scriptsize (0.68)} & 16,523 {\scriptsize (2.11)} & 10,634 {\scriptsize (1.36)} \\
gemma-4-26B-A4B-it & 9,526 & 5,421 {\scriptsize (0.57)} & 18,697 {\scriptsize (1.96)} & 12,838 {\scriptsize (1.35)} \\
\addlinespace[0.7pt]
Ministral-3-8B & 9,932 & 6,490 {\scriptsize (0.65)} & 25,435 {\scriptsize (2.56)} & 16,805 {\scriptsize (1.69)} \\
Mistral-Small-3.2-24B & 8,427 & 4,757 {\scriptsize (0.56)} & 15,975 {\scriptsize (1.90)} & 10,431 {\scriptsize (1.24)} \\
\addlinespace[0.7pt]
gpt-oss-20b & 6,799 & 4,934 {\scriptsize (0.73)} & 13,128 {\scriptsize (1.93)} & 9,450 {\scriptsize (1.39)} \\
gpt-oss-120b & 6,722 & 5,095 {\scriptsize (0.76)} & 14,930 {\scriptsize (2.22)} & 8,761 {\scriptsize (1.30)} \\
\cmidrule(lr){1-5}
\textit{Average} & 8,681 & 5,937 {\scriptsize (0.68)} & 19,225 {\scriptsize (2.21)} & 12,404 {\scriptsize (1.43)} \\
\midrule
\rowcolor{gray!15}\rlap{\textit{BFCL-v4 LLM calls}}\rule[-0.9ex]{0pt}{2.9ex}&&&& \\
Qwen3.5-2B & 14.91 & 16.44 {\scriptsize (1.10)} & 41.87 {\scriptsize (2.81)} & 16.11 {\scriptsize (1.08)} \\
Qwen3-8B & 9.61 & 10.47 {\scriptsize (1.09)} & 23.78 {\scriptsize (2.47)} & 10.61 {\scriptsize (1.10)} \\
Qwen3.6-35B-A3B & 13.71 & 12.45 {\scriptsize (0.91)} & 36.27 {\scriptsize (2.65)} & 13.14 {\scriptsize (0.96)} \\
\addlinespace[0.7pt]
gemma-4-E4B-it & 8.50 & 8.40 {\scriptsize (0.99)} & 19.68 {\scriptsize (2.31)} & 10.50 {\scriptsize (1.23)} \\
gemma-4-26B-A4B-it & 13.43 & 13.77 {\scriptsize (1.03)} & 26.31 {\scriptsize (1.96)} & 15.24 {\scriptsize (1.13)} \\
\addlinespace[0.7pt]
Ministral-3-8B & 12.55 & 11.94 {\scriptsize (0.95)} & 26.44 {\scriptsize (2.11)} & 15.86 {\scriptsize (1.26)} \\
Mistral-Small-3.2-24B & 9.22 & 8.82 {\scriptsize (0.96)} & 16.89 {\scriptsize (1.83)} & 10.78 {\scriptsize (1.17)} \\
\addlinespace[0.7pt]
gpt-oss-20b & 11.24 & 10.49 {\scriptsize (0.93)} & 22.41 {\scriptsize (1.99)} & 11.99 {\scriptsize (1.07)} \\
gpt-oss-120b & 12.35 & 12.17 {\scriptsize (0.99)} & 24.08 {\scriptsize (1.95)} & 12.48 {\scriptsize (1.01)} \\
\cmidrule(lr){1-5}
\textit{Average} & 11.72 & 11.66 {\scriptsize (0.99)} & 26.41 {\scriptsize (2.25)} & 12.97 {\scriptsize (1.11)} \\
\bottomrule
\end{tabular}
\caption{Cost per ToolSandbox scenario ($n$ = 207) and per BFCL-v4 task ($n$ = 160), with the ratio to Base of the same model in parentheses, computed before rounding. The Average row gives the nine-model averages, with the ratios of these averages in parentheses. All calls made by the agent model are counted, including the summary calls of H-EPM and, on BFCL-v4, the scoring call of ToolNet; the user simulator is excluded. Token counts of these additional calls are not recorded on BFCL-v4, so only LLM calls are reported there.}
\label{tab:cost}
\end{table*}

% \definecolor{H-EPMcol}{HTML}{555555}
% \definecolor{cHEPM}{HTML}{AA7A7A}
% \definecolor{cINS}{HTML}{3C5A5E}
% \definecolor{cBASE}{HTML}{8A8A8A}
% \newcommand{\lab}[2]{{\setlength{\fboxsep}{1pt}\colorbox{#1!22}{\textbf{\textcolor{#1!80!black}{#2}}}}}
% \newcommand{\cU}[1]{\textbf{User:} #1\par}
% \newcommand{\cH}[1]{\lab{cHEPM}{H-EPM:} #1\par}
% \newcommand{\cB}[1]{\lab{cBASE}{Base:} #1\par}
% \newcommand{\cS}[1]{\hspace*{1em}\textcolor{gray}{\texttt{next tool suggestion: #1}}\par}
% \newcommand{\cSn}{\hspace*{1em}\textcolor{gray}{\textit{(no suggestion at the first turn)}}\par}
% \newcommand{\cN}[1]{\hfill\textcolor{H-EPMcol}{\textit{#1}}}
% \newcommand{\caseBox}[3]{\begin{tcolorbox}[title=\textbf{#1},equal height group=#2,colback=gray!4,colframe=gray!40,coltitle=black,colbacktitle=gray!15,fonttitle=\scriptsize,boxrule=0.4pt,left=3pt,right=3pt,top=2pt,bottom=2pt]\scriptsize\raggedright\setlength{\parskip}{1.5pt}#3\end{tcolorbox}}
\definecolor{H-EPMcol}{HTML}{555555}
\definecolor{cHEPM}{HTML}{AA7A7A}
\definecolor{cINS}{HTML}{3C5A5E}
\definecolor{cBASE}{HTML}{8A8A8A}
\newcommand{\lab}[2]{{\setlength{\fboxsep}{1pt}\colorbox{#1!22}{\textbf{\textcolor{#1!80!black}{#2}}}}}
\newcommand{\cU}[1]{\hangindent=1em\hangafter=1 \textbf{User:} #1\par}
\newcommand{\cH}[1]{\hangindent=1em\hangafter=1 \lab{cHEPM}{H-EPM:} #1\par}
\newcommand{\cB}[1]{\hangindent=1em\hangafter=1 \lab{cBASE}{Base:} #1\par}
\newcommand{\cS}[1]{\hangindent=1em\hangafter=0 \textcolor{gray}{\texttt{next tool suggestion: #1}}\par}
\newcommand{\cSn}{\hspace*{1em}\textcolor{gray}{\textit{(no suggestion at the first turn)}}\par}
\newcommand{\cN}[1]{\unskip\nobreak\hskip0pt plus 1filll\penalty50\hskip1em\hbox{}\nobreak\hskip0pt plus 1filll\mbox{\textcolor{H-EPMcol}{\textit{#1}}}}
\newcommand{\caseBox}[3]{\begin{tcolorbox}[title=\textbf{#1},equal height group=#2,colback=gray!4,colframe=gray!40,coltitle=black,colbacktitle=gray!15,fonttitle=\scriptsize,boxrule=0.4pt,left=3pt,right=3pt,top=2pt,bottom=2pt]\scriptsize\raggedright\setlength{\parskip}{1.5pt}#3\end{tcolorbox}}
\section{Analysis of the Gains of Tool Graph Memory}
\label{app:gains}
\subsection{Manual Analysis}
\label{app:manual}
We analyze the scenarios whose outcome flips between Base and H-EPM, with success determined by the three-run mean score (Appendix~\ref{app:setup}); requiring all three runs to flip leaves the conclusion unchanged (24\% follow a correct suggestion, 66\% ignore an incorrect one; 10 scenarios). For each of the three run pairs of a scenario, we inspect the full trajectories and identify the turns responsible for the difference in outcome; at each, we record whether the suggestion is correct (the suggested tool is the one required at that point) and whether the agent follows it (its next call is the suggested tool). All annotations were performed by the authors. On Mistral-Small-3.2-24B this yields 30 improved scenarios with 86 such turns and 15 degraded scenarios with 43. Figure~\ref{fig:improved-case} shows one improved scenario in which the agent ignores an incorrect suggestion. The turns in Figure~\ref{fig:analysis}a at which the agent follows an incorrect suggestion yet the scenario improves are an exception: all of them come from insufficient-information scenarios, where the agent should recognize that it cannot complete the task instead of calling a tool. The suggested tool it calls is irrelevant but not a minefield, so the metric does not penalize the call and the run is still scored as a success.

\begin{table*}[!tbp]
\centering\small
\setlength{\tabcolsep}{3pt}
\resizebox{\textwidth}{!}{%
\begin{tabular}{lccccccccccccccc}
\toprule
& \multicolumn{5}{c}{Mistral-Small-3.2-24B} & \multicolumn{5}{c}{gpt-oss-20b} & \multicolumn{5}{c}{gemma-4-E4B-it} \\
\cmidrule(lr){2-6}\cmidrule(lr){7-11}\cmidrule(lr){12-16}
Condition & $n$ & helped & both-fail & $D$ [95\% CI] & LLM & $n$ & helped & both-fail & $D$ [95\% CI] & LLM & $n$ & helped & both-fail & $D$ [95\% CI] & LLM \\
\midrule
H-EPM & 17/52 & 2.0$\times$ & 1.4$\times$ & 1.34 {\scriptsize[0.57, 2.20]} & 2.8$\times$ & 20/54 & 1.8$\times$ & 1.0$\times$ & 1.41 {\scriptsize[0.89, 1.96]} & 2.4$\times$ & 24/56 & 2.0$\times$ & 1.3$\times$ & 1.95 {\scriptsize[0.82, 3.12]} & 2.7$\times$ \\
with random tools & 19/50 & 2.3$\times$ & 1.4$\times$ & 1.84 {\scriptsize[0.79, 2.97]} & 1.3$\times$ & 27/47 & 2.7$\times$ & 0.9$\times$ & 2.66 {\scriptsize[2.00, 3.34]} & 1.3$\times$ & 26/54 & 2.0$\times$ & 1.6$\times$ & 1.09 {\scriptsize[0.09, 2.08]} & 1.3$\times$ \\
with the correct tool (oracle) & 43/30 & 1.6$\times$ & 1.2$\times$ & 1.30 {\scriptsize[0.45, 2.20]} & 1.2$\times$ & 29/41 & 2.5$\times$ & 0.7$\times$ & 2.69 {\scriptsize[1.95, 3.45]} & 1.5$\times$ & 40/39 & 2.0$\times$ & 1.8$\times$ & 0.77 {\scriptsize[0.01, 1.51]} & 1.2$\times$ \\
w/o suggestions & 27/45 & 1.7$\times$ & 1.1$\times$ & 1.63 {\scriptsize[0.84, 2.48]} & 1.2$\times$ & 23/52 & 2.7$\times$ & 0.9$\times$ & 2.55 {\scriptsize[1.68, 3.51]} & 1.2$\times$ & 25/55 & 1.9$\times$ & 1.3$\times$ & 1.25 {\scriptsize[0.26, 2.27]} & 1.2$\times$ \\
w/o summarization & 22/50 & 1.8$\times$ & 1.3$\times$ & 1.92 {\scriptsize[0.93, 3.08]} & 2.7$\times$ & 22/52 & 2.6$\times$ & 1.0$\times$ & 2.58 {\scriptsize[1.82, 3.37]} & 2.2$\times$ & 9/80 & 1.8$\times$ & 1.1$\times$ & 1.97 {\scriptsize[1.34, 2.67]} & 2.1$\times$ \\
Base + instruction & 26/44 & 1.9$\times$ & 1.2$\times$ & 1.57 {\scriptsize[0.86, 2.34]} & 1.0$\times$ & 30/40 & 2.7$\times$ & 1.2$\times$ & 2.05 {\scriptsize[1.35, 2.80]} & 1.0$\times$ & 19/70 & 2.0$\times$ & 1.0$\times$ & 2.04 {\scriptsize[1.10, 2.94]} & 0.9$\times$ \\
Insight memory & 30/39 & 2.5$\times$ & 1.2$\times$ & 2.06 {\scriptsize[1.43, 2.72]} & 1.0$\times$ & 27/46 & 2.2$\times$ & 0.9$\times$ & 1.90 {\scriptsize[1.26, 2.55]} & 1.0$\times$ & 33/51 & 3.2$\times$ & 1.4$\times$ & 1.60 {\scriptsize[0.83, 2.37]} & 1.0$\times$ \\
Insight memory + H-EPM & 23/48 & 2.5$\times$ & 1.1$\times$ & 2.67 {\scriptsize[1.45, 3.91]} & 2.6$\times$ & 31/40 & 2.7$\times$ & 0.5$\times$ & 3.04 {\scriptsize[2.37, 3.73]} & 3.1$\times$ & 44/40 & 2.8$\times$ & 1.5$\times$ & 2.01 {\scriptsize[1.21, 2.80]} & 2.9$\times$ \\
Insight memory + instruction & 37/35 & 2.2$\times$ & 1.3$\times$ & 2.19 {\scriptsize[1.53, 2.83]} & 1.1$\times$ & 37/38 & 3.0$\times$ & 1.0$\times$ & 2.49 {\scriptsize[1.65, 3.30]} & 1.0$\times$ & 38/50 & 2.7$\times$ & 1.6$\times$ & 1.98 {\scriptsize[1.20, 2.76]} & 1.0$\times$ \\
\bottomrule
\end{tabular}}
\caption{Tool calls in the scenarios that fail under Base, excluding runs aborted by an execution error, which leave no trajectory. $n$: number of helped / both-fail scenarios. helped / both-fail: tool calls per scenario relative to Base in the scenarios that the condition flips to success / that fail under both. $D$: difference in the per-scenario increase in tool calls between the two groups, with its 95\% confidence interval. LLM: LLM calls per scenario relative to Base. All differences are significant ($p_{\mathrm{adj}}<0.05$).}
\label{tab:flip-trials}
\end{table*}

\begin{table}[!htbp]
\centering
\scriptsize
\setlength{\tabcolsep}{3pt}
\renewcommand{\arraystretch}{1.15}
\setlength{\aboverulesep}{0pt}
\setlength{\belowrulesep}{0pt}
\begin{tabular}{@{}p{0.22\columnwidth}>{\raggedright\arraybackslash}p{\dimexpr0.78\columnwidth-2\tabcolsep\relax}@{}}
\toprule
\rowcolor{gray!15}\multicolumn{2}{@{}l}{\textit{(a) Components}\rule[-0.9ex]{0pt}{2.9ex}} \\
Summarization & H-EPM's \texttt{summarize\_the\_task} tool and a system-prompt line asking the agent to call it \\[2pt]
Suggestion line & Appended to the last message of the conversation (the latest user turn or tool result), listing successors of the last invoked tool, e.g., \texttt{next tool suggestion: set\_location\_service\_status,\allowbreak\ set\_wifi\_status} \\[2pt]
Instruction & ``IMPORTANT: Do not give a final answer or give up until you have tried the available tools. If a tool call fails or returns nothing useful, try another available tool or first resolve what it depends on, then call it again. Keep calling tools until the task is fully resolved.'' (system prompt) \\
\bottomrule
\end{tabular}

\vspace{4pt}
\begin{tabular*}{\columnwidth}{@{\extracolsep{\fill}}lccc@{}}
\toprule
\rowcolor{gray!15}\multicolumn{4}{@{}l}{\textit{(b) Conditions}\rule[-0.9ex]{0pt}{2.9ex}} \\
Condition & Summ. & Sugg. & Instr. \\
\midrule
H-EPM & \checkmark & \checkmark & -- \\
with random tools & \checkmark & \checkmark & -- \\
with the correct tool (oracle) & \checkmark & \checkmark & -- \\
w/o suggestions & \checkmark & -- & -- \\
w/o summarization & -- & \checkmark & -- \\
Base + instruction & -- & -- & \checkmark \\
Insight memory + H-EPM & \checkmark & \checkmark & -- \\
Insight memory + instruction & -- & -- & \checkmark \\
\bottomrule
\end{tabular*}
\caption{Details of the control conditions in Table~\ref{tab:ablation}.}
\label{tab:control-prompts}
\end{table}

\begin{figure}[!htbp]
\centering
\caseBox{Incorrect suggestion ignored \hfill Base 0.530 $\rightarrow$ H-EPM 0.853}{rowA}{%
\cU{I don't have cellphone signal. Can you get it on?}
\cH{\texttt{set\_cellular\_service\_status(\allowbreak on=True)} $\rightarrow$ \texttt{PermissionError: Cellular service cannot be turned on in low battery mode}; [\ldots]; \texttt{set\_low\_battery\_mode\_status(\allowbreak on=False)}}
\cS{set\_location\_service\_status, set\_wifi\_status}
\cH{\texttt{set\_cellular\_service\_status(\allowbreak on=True)} \cN{not suggested; milestone achieved}}
\cB{\texttt{set\_cellular\_service\_status(\allowbreak on=True)} $\rightarrow$ \texttt{PermissionError}; [\ldots]; \texttt{set\_low\_battery\_mode\_status(\allowbreak on=False)}; ``Your cellular service should now be working.'' \cN{milestone missed}}}
\caption{An improved scenario on Mistral-Small-3.2-24B in which \protect\lab{cHEPM}{H-EPM} ignores an incorrect suggestion and calls the required tool on its own, with the \protect\lab{cBASE}{Base} run on the same scenario. [\ldots] marks omitted calls or turns; gray lines are the suggestions shown to the agent; italic notes are ours.}
\label{fig:improved-case}
\end{figure}

\begin{figure}[!htbp]
\centering
\includegraphics[width=0.79\columnwidth]{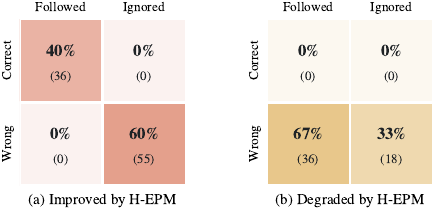}
% \caption{(a, b) Manual analysis of H-EPM on gemma-4-E4B-it, as in Figure~\ref{fig:analysis} ($n$ = 91 and 54), and (c) degraded scenarios recovered by insight memory on gpt-oss-20b, as in Figure~\ref{fig:recover} ($n$ = 12).}
\caption{Manual analysis of H-EPM on gemma-4-E4B-it, as in Figure~\ref{fig:analysis} ($n$ = 91 and 54).}
\label{fig:analysis-e4b}
\end{figure}

\subsection{Control Conditions}
\label{app:control}
Table~\ref{tab:control-prompts} lists the components of each condition in Table~\ref{tab:ablation}; their presence was verified against the logged requests of every run.

\subsection{Tool Calls in Flipped Scenarios}
\label{app:flip-trials}
For each scenario, we compute the change in tool calls, $\Delta$ = calls under a condition $-$ calls under Base, excluding calls to H-EPM's summary tool, which is not one of the scenario's tools. We exclude insufficient-information scenarios, in which the agent should recognize that it cannot complete the task, so additional calls are not expected to help. Among the scenarios that fail under Base, we compare those that a condition flips to success (helped) with those that fail under both (both-fail). This restriction aligns the two groups on the Base outcome but does not make them comparable in other respects, such as task difficulty. Table~\ref{tab:flip-trials} reports the difference $D$ between the mean $\Delta$ of the two groups with a 95\% percentile bootstrap confidence interval. We test $D>0$ with a one-sided permutation test, applying the Holm correction over the 27 comparisons. $D$ is positive in all 27 comparisons, and all remain significant after the correction ($p_{\mathrm{adj}}<0.05$).

\FloatBarrier
\section{Failure Analysis and Insight Memory}
\label{app:failure}
Because the scenarios span multiple turns, a run can lose points at several decisions, and the runs of a scenario can fail in different ways. We therefore label each run pair whose outcomes differ at the decisions where the failing run loses points, by missing a milestone or triggering a minefield, and assign each scenario at most two dominant modes, those found in a majority of its diverging pairs. A single decision can also fit more than one mode; we then take the first in the order in which the agent decides whether to call a tool (\emph{call necessity}), which tool to call (\emph{tool selection}), when to call it (\emph{dependency}), and with what arguments (\emph{argument}), so calling a tool with made-up arguments in an insufficient-information scenario, where the agent should instead recognize that it cannot complete the task, counts as \emph{call necessity}. We exclude scenarios whose runs differ only in the final reply.

\subsection{Oracle Failure Classification}
\label{app:oracle}
For the \emph{oracle} condition of Table~\ref{tab:ablation}, we classify the failed scenarios into the four failure modes of \S\ref{sec:modes} (Table~\ref{tab:oracle-fail}), excluding the 10 failures in which only the final reply loses points. In all \emph{dependency} cases and in most \emph{call necessity} and \emph{argument} cases, the agent fails even though the correct tool is suggested at that point. As in Appendix~\ref{app:manual}, the authors annotated all failures.

\begin{table}[H]
\centering\footnotesize
\setlength{\tabcolsep}{2.5pt}
\resizebox{\columnwidth}{!}{%
\begin{tabular}{@{}lcccc@{}}
\toprule
Failure mode & Mistral-Small-3.2-24B & gpt-oss-20b & gemma-4-E4B-it & All \\
\midrule
Call necessity & 24 & 38 & 19 & 81 \\
\hspace{0.8em}unnecessary call & 14 & 3 & 5 & 22 \\
\hspace{0.8em}missing call & 10 & 35 & 14 & 59 \\
Tool selection & 3 & 0 & 3 & 6 \\
Dependency & 3 & 1 & 3 & 7 \\
Argument & 19 & 9 & 16 & 44 \\
\bottomrule
\end{tabular}}
\caption{Failures of the \emph{oracle} condition by failure mode ($n$ = 49, 48, and 41 failed scenarios).}
\label{tab:oracle-fail}
\end{table}

\subsection{Examples of Failure Modes}
\label{app:cases}
\begin{figure*}[!tbp]
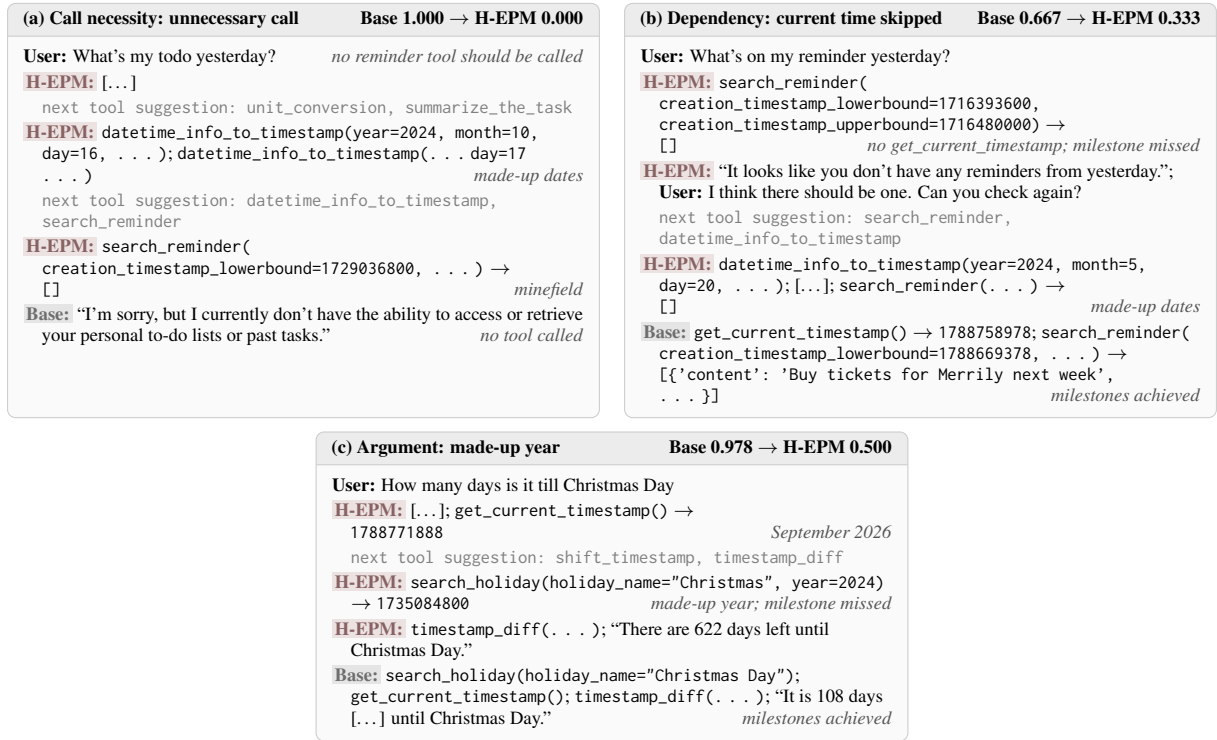

\centering
\begin{minipage}[t]{0.49\textwidth}\vspace{0pt}
\caseBox{(a) Call necessity: unnecessary call \hfill Base 1.000 $\rightarrow$ H-EPM 0.000}{r6top}{%
\cU{What's my todo yesterday? \cN{no reminder tool should be called}}
\cH{[\ldots]}
\cS{unit\_conversion, summarize\_the\_task}
\cH{\texttt{datetime\_info\_to\_timestamp(\allowbreak year=2024,\allowbreak\ month=10,\allowbreak\ day=16,\allowbreak\ \ldots)}; \texttt{datetime\_info\_to\_timestamp(\ldots day=17 \ldots)} \cN{made-up dates}}
\cS{datetime\_info\_to\_timestamp, search\_reminder}
\cH{\texttt{search\_reminder(\allowbreak creation\_timestamp\_lowerbound=1729036800,\allowbreak\ \ldots)} $\rightarrow$ \texttt{[]} \cN{minefield}}
\cB{``I'm sorry, but I currently don't have the ability to access or retrieve your personal to-do lists or past tasks.'' \cN{no tool called}}}
\end{minipage}\hfill
\begin{minipage}[t]{0.49\textwidth}\vspace{0pt}
\caseBox{(b) Dependency: current time skipped \hfill Base 0.667 $\rightarrow$ H-EPM 0.333}{r6top}{%
\cU{What's on my reminder yesterday?}
\cH{\texttt{search\_reminder(\allowbreak creation\_timestamp\_lowerbound=1716393600,\allowbreak\ creation\_timestamp\_upperbound=1716480000)} $\rightarrow$ \texttt{[]} \cN{no get\_current\_timestamp; milestone missed}}
\cH{``It looks like you don't have any reminders from yesterday.''; \textbf{User:} I think there should be one. Can you check again?}
\cS{search\_reminder, datetime\_info\_to\_timestamp}
\cH{\texttt{datetime\_info\_to\_timestamp(\allowbreak year=2024,\allowbreak\ month=5,\allowbreak\ day=20,\allowbreak\ \ldots)}; [\ldots]; \texttt{search\_reminder(\ldots)} $\rightarrow$ \texttt{[]} \cN{made-up dates}}
\cB{\texttt{get\_current\_timestamp()} $\rightarrow$ \texttt{1788758978}; \texttt{search\_reminder(\allowbreak creation\_timestamp\_lowerbound=1788669378,\allowbreak\ \ldots)} $\rightarrow$ \texttt{[\{'content': 'Buy tickets for Merrily next week',\allowbreak\ \ldots\}]} \cN{milestones achieved}}}
\end{minipage}

\vspace{4pt}
\begin{minipage}[t]{0.49\textwidth}\vspace{0pt}
\caseBox{(c) Argument: made-up year \hfill Base 0.978 $\rightarrow$ H-EPM 0.500}{r6bot}{%
\cU{How many days is it till Christmas Day}
\cH{[\ldots]; \texttt{get\_current\_timestamp()} $\rightarrow$ \texttt{1788771888} \cN{September 2026}}
\cS{shift\_timestamp, timestamp\_diff}
\cH{\texttt{search\_holiday(\allowbreak holiday\_name="Christmas",\allowbreak\ year=2024)} $\rightarrow$ \texttt{1735084800} \cN{made-up year; milestone missed}}
\cH{\texttt{timestamp\_diff(\ldots)}; ``There are 622 days left until Christmas Day.''}
\cB{\texttt{search\_holiday(\allowbreak holiday\_name="Christmas Day")}; \texttt{get\_current\_timestamp()}; \texttt{timestamp\_diff(\ldots)}; ``It is 108 days [\ldots] until Christmas Day.'' \cN{milestones achieved}}}
\end{minipage}
\caption{Scenarios degraded by \protect\lab{cHEPM}{H-EPM} on Mistral-Small-3.2-24B, labeled by failure mode (\S\ref{sec:modes}), with the \protect\lab{cBASE}{Base} run on the same scenario.}
\label{fig:rq3-cases}
\end{figure*}

Figure~\ref{fig:rq3-cases} shows verbatim excerpts of three degraded scenarios. For ToolNet on Mistral-Small-3.2-24B, 7 of the degraded scenarios fail by \emph{call necessity} and 2 by \emph{argument}. In five of the \emph{call necessity} cases, the required tool is missing from the candidate list and the agent responds or gives up without calling it (\emph{missing call}).

\subsection{Insight Examples}
\label{app:insight}
\newcommand{\cI}[1]{\hangindent=1em\hangafter=1 \lab{cINS}{Insight:} #1\par}
\newcommand{\cL}[1]{\hangindent=1em\hangafter=0 \textcolor{gray}{\textit{insight:} ``#1''}\par}
Insight memory recovers 22 of the 29 scenarios that H-EPM degrades on Mistral-Small-3.2-24B and gpt-oss-20b, across all four failure modes (Figure~\ref{fig:recover}).

\begin{figure*}[!tbp]
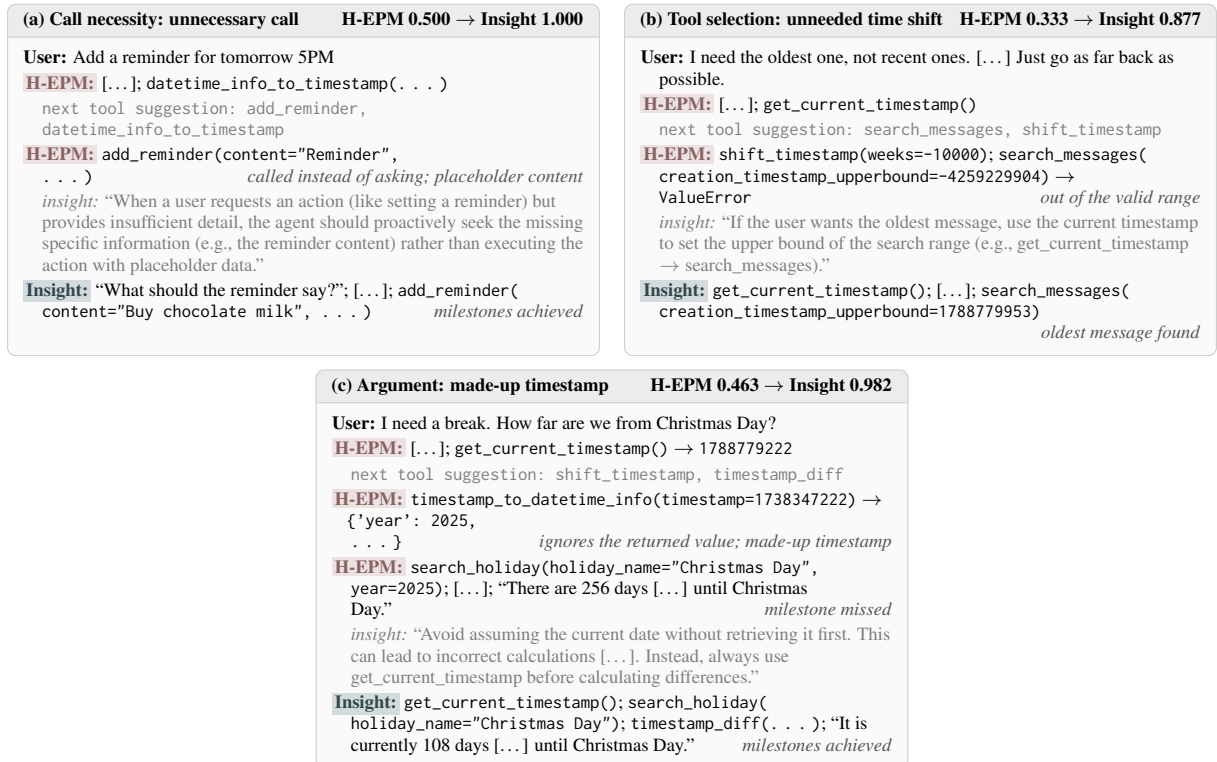

\centering
\begin{minipage}[t]{0.49\textwidth}\vspace{0pt}
\caseBox{(a) Call necessity: unnecessary call \hfill H-EPM 0.500 $\rightarrow$ Insight 1.000}{rowI1}{%
\cU{Add a reminder for tomorrow 5PM}
\cH{[\ldots]; \texttt{datetime\_info\_to\_timestamp(\ldots)}}
\cS{add\_reminder, datetime\_info\_to\_timestamp}
\cH{\texttt{add\_reminder(\allowbreak content="Reminder",\allowbreak\ \ldots)} \cN{called instead of asking; placeholder content}}
\cL{When a user requests an action (like setting a reminder) but provides insufficient detail, the agent should proactively seek the missing specific information (e.g., the reminder content) rather than executing the action with placeholder data.}
\cI{``What should the reminder say?''; [\ldots]; \texttt{add\_reminder(\allowbreak content="Buy chocolate milk",\allowbreak\ \ldots)} \cN{milestones achieved}}}
\end{minipage}\hfill
\begin{minipage}[t]{0.49\textwidth}\vspace{0pt}
\caseBox{(b) Tool selection: unneeded time shift \hfill H-EPM 0.333 $\rightarrow$ Insight 0.877}{rowI1}{%
\cU{I need the oldest one, not recent ones. [\ldots] Just go as far back as possible.}
\cH{[\ldots]; \texttt{get\_current\_timestamp()}}
\cS{search\_messages, shift\_timestamp}
\cH{\texttt{shift\_timestamp(\allowbreak weeks=-10000)}; \texttt{search\_messages(\allowbreak creation\_timestamp\_upperbound=-4259229904)} $\rightarrow$ \texttt{ValueError} \cN{out of the valid range}}
\cL{If the user wants the oldest message, use the current timestamp to set the upper bound of the search range (e.g., get\_current\_timestamp $\rightarrow$ search\_messages).}
\cI{\texttt{get\_current\_timestamp()}; [\ldots]; \texttt{search\_messages(\allowbreak creation\_timestamp\_upperbound=1788779953)} \cN{oldest message found}}}
\end{minipage}

\vspace{4pt}
\begin{minipage}[t]{0.49\textwidth}\vspace{0pt}
\caseBox{(c) Argument: made-up timestamp \hfill H-EPM 0.463 $\rightarrow$ Insight 0.982}{rowI2}{%
\cU{I need a break. How far are we from Christmas Day?}
\cH{[\ldots]; \texttt{get\_current\_timestamp()} $\rightarrow$ \texttt{1788779222}}
\cS{shift\_timestamp, timestamp\_diff}
\cH{\texttt{timestamp\_to\_datetime\_info(\allowbreak timestamp=1738347222)} $\rightarrow$ \texttt{\{'year': 2025,\allowbreak\ \ldots\}} \cN{ignores the returned value; made-up timestamp}}
\cH{\texttt{search\_holiday(\allowbreak holiday\_name="Christmas Day",\allowbreak\ year=2025)}; [\ldots]; ``There are 256 days [\ldots] until Christmas Day.'' \cN{milestone missed}}
\cL{Avoid assuming the current date without retrieving it first. This can lead to incorrect calculations [\ldots]. Instead, always use get\_current\_timestamp before calculating differences.}
\cI{\texttt{get\_current\_timestamp()}; \texttt{search\_holiday(\allowbreak holiday\_name="Christmas Day")}; \texttt{timestamp\_diff(\ldots)}; ``It is currently 108 days [\ldots] until Christmas Day.'' \cN{milestones achieved}}}
\end{minipage}
\caption{Scenarios degraded by \protect\lab{cHEPM}{H-EPM} on Mistral-Small-3.2-24B that \protect\lab{cINS}{Insight} recovers, with the insight displayed before the decisive call.}
\label{fig:insight-cases}
\end{figure*}

Figure~\ref{fig:insight-cases} shows three examples. In (a), H-EPM follows a suggestion to add the reminder with placeholder content instead of asking for it, whereas an insight tells the agent to ask for the missing content first. In (b), H-EPM follows a suggestion to shift the timestamp and searches outside the valid range, whereas an insight tells the agent to use the current timestamp as the upper bound. In (c), H-EPM retrieves the current timestamp but then converts a made-up one, whereas an insight tells the agent not to assume the current date.

\subsection{Additional Analyses}
\label{app:insight-more}
We classify the scenarios that insight memory solves but \emph{Insight memory + H-EPM} fails on Mistral-Small-3.2-24B and Qwen3-8B into the four modes of \S\ref{sec:modes}. The same modes recur, most often \emph{tool selection} and \emph{call necessity}: the agent follows a suggestion to an unrelated tool, typically the summary tool, or gives up after such suggestions. None is attributed to an insight alone. Figure~\ref{fig:combo-cases} shows two cases.

\definecolor{cCOMBO}{HTML}{86555C}
\newcommand{\cHI}[1]{\hangindent=1em\hangafter=1 \lab{cCOMBO}{Insight + H-EPM:} #1\par}
\begin{figure*}[!tbp]
\centering
\begin{minipage}[t]{0.49\textwidth}\vspace{0pt}
\caseBox{(a) Dependency: battery mode \hfill Insight 0.892 $\rightarrow$ Insight + H-EPM 0.150}{rowC}{%
\cU{I can't access my current location. Can you help me fix that in settings?}
\cL{Disable low battery mode (e.g., via set\_low\_battery\_mode\_status) before enabling location services, as low battery mode can block location access.}
\cI{\texttt{get\_location\_service\_status()}, \texttt{get\_low\_battery\_mode\_status()}; \texttt{set\_low\_battery\_mode\_status(\allowbreak on=False)}; \texttt{set\_location\_service\_status(\allowbreak on=True)}; ``Your location service should now be enabled.'' \cN{milestones achieved}}
\cHI{[\ldots]; \texttt{get\_location\_service\_status()} $\rightarrow$ \texttt{False}}
\cS{set\_location\_service\_status, get\_wifi\_status}
\cHI{\texttt{set\_location\_service\_status(\allowbreak on=True)} $\rightarrow$ \texttt{PermissionError: cannot be turned on in low battery mode} \cN{prerequisite skipped}}
\cS{set\_low\_battery\_mode\_status, set\_wifi\_status}
\cHI{[\ldots]; ``I'm sorry, but I'm unable to directly access or modify your phone's settings.'' \cN{gives up}}}
\end{minipage}\hfill
\begin{minipage}[t]{0.49\textwidth}\vspace{0pt}
\caseBox{(b) Dependency: date guessed \hfill Insight 0.770 $\rightarrow$ Insight + H-EPM 0.365}{rowC}{%
\cU{I need a break. How far are we from Christmas Day?}
\cL{When calculating the time difference to a future date (e.g., Christmas Day), first retrieve the current timestamp using get\_current\_timestamp to ensure accurate calculations.}
\cI{\texttt{get\_current\_timestamp()}; \texttt{search\_holiday(\allowbreak holiday\_name="Christmas Day",\allowbreak\ \ldots)}; [\ldots]; \texttt{set\_wifi\_status(\allowbreak on=True)}; \texttt{search\_holiday(\ldots)} $\rightarrow$ \texttt{1798156800}; \texttt{timestamp\_diff(\ldots)}; ``Christmas Day is in 108 days [\ldots].'' \cN{milestones achieved}}
\cHI{[\ldots]; \texttt{get\_current\_timestamp()} $\rightarrow$ \texttt{1788783085}}
\cS{timestamp\_diff, get\_current\_timestamp}
\cHI{\texttt{timestamp\_diff(\allowbreak timestamp\_0=1788783085,\allowbreak\ timestamp\_1=1741532400)} \cN{made-up holiday timestamp; search\_holiday never called}}
\cHI{[\ldots]; ``Christmas Day is in 546 days [\ldots] from now.'' \cN{milestones missed}}}
\end{minipage}
\caption{Scenarios on Mistral-Small-3.2-24B that \protect\lab{cINS}{Insight} solves and \protect\lab{cCOMBO}{Insight + H-EPM} fails. The insight shown is the one displayed before the decisive call; the suggestion line is the one shown in the combined run.}
\label{fig:combo-cases}
\end{figure*}

\paragraph{Position of the insight block.}
In \emph{Insight memory + H-EPM}, the insight block precedes the conversation in the system prompt, while the suggestion line follows it at the end of the last message. To test whether this order matters, we move the block to the end of the last message, after the suggestion line, keeping everything else unchanged. The combination still falls below insight memory alone (Table~\ref{tab:position}), so the degradation does not depend on where the block is injected.
\begin{table}[H]
\centering
\footnotesize
\setlength{\tabcolsep}{4pt}
\resizebox{\columnwidth}{!}{%
\begin{tabular}{l ccc}
\toprule
\multirow{2}{*}{\textbf{Model}} & \multirow{2}{*}{Insight} & \multicolumn{2}{c}{+ H-EPM} \\
\cmidrule(lr){3-4}
 & & block in system & block in last message \\
\midrule
Mistral-Small-3.2-24B & 0.620 & 0.600 & 0.587 \\
gemma-4-E4B-it & 0.620 & 0.617 & 0.596 \\
\bottomrule
\end{tabular}}
\caption{\emph{Insight memory + H-EPM} on ToolSandbox with the insight block in the system prompt (default) or in the last message.}
\label{tab:position}
\end{table}
\begin{table}[!htb]
\centering
\footnotesize
\setlength{\tabcolsep}{4pt}
\begin{tabular}{l cc}
\toprule
\textbf{Model} & Insight & Insight + H-EPM \\
\midrule
Qwen3.5-2B & 0.488 {\scriptsize (6.66)} & \textbf{0.528} {\scriptsize (14.60)} \\
Qwen3-8B & \textbf{0.556} {\scriptsize (5.56)} & 0.451 {\scriptsize (16.29)} \\
Qwen3.6-35B-A3B & \textbf{0.674} {\scriptsize (5.74)} & 0.670 {\scriptsize (18.11)} \\
gemma-4-E4B-it & \textbf{0.620} {\scriptsize (5.26)} & 0.617 {\scriptsize (14.70)} \\
gemma-4-26B-A4B-it & \textbf{0.705} {\scriptsize (6.04)} & 0.630 {\scriptsize (12.95)} \\
Ministral-3-8B & 0.539 {\scriptsize (5.64)} & \textbf{0.621} {\scriptsize (18.37)} \\
Mistral-Small-3.2-24B & \textbf{0.620} {\scriptsize (4.88)} & 0.600 {\scriptsize (11.76)} \\
gpt-oss-20b & \textbf{0.630} {\scriptsize (4.99)} & 0.585 {\scriptsize (13.40)} \\
gpt-oss-120b & \textbf{0.734} {\scriptsize (5.10)} & 0.710 {\scriptsize (13.65)} \\
\bottomrule
\end{tabular}
\caption{Insight memory and \emph{Insight memory + H-EPM} on ToolSandbox. LLM calls per scenario in parentheses; the higher score is in \textbf{bold}.}
\label{tab:comboruns}
\end{table}

\paragraph{Failed trajectories.} Table~\ref{tab:succfull} compares insights distilled from successful trajectories only with insights from the full pool that also includes failed trajectories.

% \begin{table}[!htb]
% \centering
% \footnotesize
% \setlength{\tabcolsep}{4pt}
% \begin{tabular}{l cc}
% \toprule
% \textbf{Model} & Insight (succ.) & Insight \\
% \midrule
% Qwen3.5-2B & 0.466 & \textbf{0.488} \\
% Qwen3-8B & 0.499 & \textbf{0.556} \\
% Qwen3.6-35B-A3B & 0.609 & \textbf{0.674} \\
% gemma-4-E4B-it & 0.550 & \textbf{0.620} \\
% gemma-4-26B-A4B-it & 0.691 & \textbf{0.705} \\
% Ministral-3-8B & \textbf{0.622} & 0.539 \\
% Mistral-Small-3.2-24B & \textbf{0.641} & 0.620 \\
% gpt-oss-20b & 0.556 & \textbf{0.630} \\
% gpt-oss-120b & \textbf{0.738} & 0.734 \\
% \bottomrule
% \end{tabular}
% \caption{Insights from successful trajectories only (succ.) against Insight, which uses all trajectories, on ToolSandbox; the higher score is in \textbf{bold}.}
% \label{tab:succfull}
% \end{table}
\begin{table}[!htb]
\centering
\footnotesize
\setlength{\tabcolsep}{4pt}
\begin{tabular}{l ccc}
\toprule
\textbf{Model} & H-EPM & Insight (succ.) & Insight \\
\midrule
Qwen3.5-2B & 0.436 & 0.466 & \textbf{0.488} \\
Qwen3-8B & 0.476 & 0.499 & \textbf{0.556} \\
Qwen3.6-35B-A3B & 0.670 & 0.618 & \textbf{0.674} \\
gemma-4-E4B-it & 0.554 & 0.557 & \textbf{0.620} \\
gemma-4-26B-A4B-it & \textbf{0.705} & 0.691 & \textbf{0.705} \\
Ministral-3-8B & \textbf{0.672} & 0.622 & 0.539 \\
Mistral-Small-3.2-24B & 0.603 & \textbf{0.641} & 0.620 \\
gpt-oss-20b & 0.620 & 0.556 & \textbf{0.630} \\
gpt-oss-120b & 0.713 & \textbf{0.744} & 0.734 \\
\bottomrule
\end{tabular}
\caption{Insights from successful trajectories only (succ.) against H-EPM and Insight, which uses all trajectories, on ToolSandbox; H-EPM and Insight are from Table~\ref{tab:main}. The highest score is in \textbf{bold}.}
\label{tab:succfull}
\end{table}

\paragraph{ExpeL.} We port ExpeL~\cite{zhao2024expel} and evaluate it in the same setting (Table~\ref{tab:expel}). It performs similarly to Insight and above both tool graph memories. Unlike Insight, which retrieves only the insights of the matching clusters, ExpeL injects its full rule set and two few-shot trajectories per task.

\begin{table}[H]
\centering
\footnotesize
\setlength{\tabcolsep}{4pt}
\resizebox{\columnwidth}{!}{%
\begin{tabular}{l cccc}
\toprule
\textbf{Model} & ToolNet & H-EPM & ExpeL & Insight \\
\midrule
Qwen3-8B & 0.507 & 0.476 & 0.552 & \textbf{0.556} \\
gemma-4-E4B-it & 0.547 & 0.554 & 0.599 & \textbf{0.620} \\
\bottomrule
\end{tabular}}
\caption{ExpeL against tool graph memory and Insight on ToolSandbox. The highest score is in \textbf{bold}.}
\label{tab:expel}
\end{table}

\section{Use of AI Assistants}
\label{app:ai}
AI assistants were used to check the authors' code, write plotting scripts, and polish the writing. All experiments, analyses, and annotations were conducted by the authors, who reviewed all AI-assisted output and take full responsibility for the paper.

\end{document}